\documentclass[journal,10pt,final]{IEEEtran}

\pdfoutput=1

\usepackage{cite}
\usepackage{amsmath}
\usepackage{amssymb}
\usepackage{amsthm}
\usepackage{mathtools}
\usepackage{graphicx}
\usepackage{caption}
\usepackage{subcaption}
\usepackage{algorithm}
\usepackage{hyperref}
\usepackage{multirow}
\usepackage{soul}
\usepackage{color}
\usepackage{bm}
\usepackage{dsfont}
\usepackage[multiple]{footmisc}

\newcommand{\C}{\mathbb{C}}

\newcommand{\rank}{\mathbf{rank}}

\newcommand{\prox}{\mathbf{prox}}
\newcommand{\rs}{\mathcal{R}}
\newcommand{\ie}{i.e.,~}
\newcommand{\eg}{e.g.,~}

\usepackage{multibib}
\newcites{Supp}{References}
\bstctlcite[@auxoutSupp]{BSTcontrol2}

\begin{document}

\title{Online Adaptive Image Reconstruction (OnAIR) Using Dictionary Models}

\author{Brian~E.~Moore,~\IEEEmembership{Member,~IEEE,}~Saiprasad~Ravishankar,~\IEEEmembership{Member,~IEEE,} \\
Raj~Rao~Nadakuditi,~\IEEEmembership{Member,~IEEE,}~and~Jeffrey~A.~Fessler,~\IEEEmembership{Fellow,~IEEE} \vspace{-0.05in}

\thanks{\copyright~2019 IEEE. Personal use of this material is permitted. Permission from IEEE must be obtained for all other uses, in any current or future media, including reprinting/republishing this material for advertising or promotional purposes, creating new collective works, for resale or redistribution to servers or lists, or reuse of any copyrighted component of this work in other works.}

\thanks{This work was supported in part by the following grants: ONR grant N00014-15-1-2141, DARPA Young Faculty Award D14AP00086, ARO MURI grants W911NF-11-1-0391 and 2015-05174-05, NIH grants R01 EB 023618 and P01 CA 059827, and a UM-SJTU seed grant.}

\thanks{B.~E.~Moore, S.~Ravishankar, R.~R.~Nadakuditi, and J.~A.~Fessler are with the Department of Electrical Engineering and Computer Science, University of Michigan, Ann Arbor, MI, 48109 USA (email: brimoor@umich.edu, ravisha@umich.edu, rajnrao@umich.edu, fessler@umich.edu).}

\thanks{This paper has supplementary downloadable material available at \url{http://ieeexplore.ieee.org}, provided by the author. The material includes additional experimental results that provide insights into the properties of the methods proposed in this work. The material is 400KB in size.}
}

\markboth{IEEE TRANSACTIONS ON COMPUTATIONAL IMAGING}{}

\maketitle

\IEEEpeerreviewmaketitle

\begin{abstract}
Sparsity and low-rank models have been popular for reconstructing images and videos from limited or corrupted measurements. Dictionary or transform learning methods are useful in applications such as denoising, inpainting, and medical image reconstruction. This paper proposes a framework for online (or time-sequential) adaptive reconstruction of dynamic image sequences from linear (typically undersampled) measurements. We model the spatiotemporal patches of the underlying dynamic image sequence as sparse in a dictionary, and we simultaneously estimate the dictionary and the images sequentially from streaming measurements. Multiple constraints on the adapted dictionary are also considered such as a unitary matrix, or low-rank dictionary atoms that provide additional efficiency or robustness. The proposed online algorithms are memory efficient and involve simple updates of the dictionary atoms, sparse coefficients, and images. Numerical experiments demonstrate the usefulness of the proposed methods in inverse problems such as video reconstruction or inpainting from noisy, subsampled pixels, and dynamic magnetic resonance image reconstruction from very limited measurements.

\end{abstract}



\section{Introduction} \label{sec:intro}

Models of signals and images based on sparsity, low-rank, and other properties are useful in image and video processing.
In ill-posed or ill-conditioned inverse problems, it is often useful to employ signal models that reflect known or assumed properties of the latent images. Such models are often used to construct appropriate regularization. For example, the sparsity of images in wavelet or discrete cosine transform (DCT) domains  has been exploited for image and video reconstruction tasks \cite{rajpoot1,Rusanovskyy2005, lustig}.
In particular, the learning of such models has been explored in various settings \cite{dl1:elad, Yagh, sabres, saibres2222,wensabres556, elad3, bresai}, where they may potentially outperform fixed models since they adapt to signals and signal classes.


There has 
been growing interest in such dictionary learning-based 
image restoration or reconstruction methods \cite{elad4, lingal16, saibriarajjeff2}. For example, in blind compressed sensing \cite{bresai,lingal16, saibriarajjeff2}, a dictionary for the underlying image or video is estimated together with the image from undersampled measurements. This allows the dictionary to adapt to the current data, which may enable learning novel features, providing improved reconstructions. 

For inverse problems involving large-scale or streaming data, \eg in interventional imaging or restoring (e.g, denoising, inpainting) large or streaming videos, etc., it is often critical to obtain reconstructions in an online or time-sequential (or data-sequential) manner to limit latency. Batch methods that process all the data at once are often prohibitively expensive in terms of time and memory usage, and infeasible for streaming data. 
Methods for online learning of dictionary and sparsifying transform models from streaming signals including noisy data have been recently proposed \cite{Mai, saonli1, saonli2, wensaividosat, wen2018vidosat} and shown to outperform state-of-the-art methods \cite{guo2016video, sutour2014adaptive, zhang2017beyond} including deep learning methods for video denoising.

This paper focuses on methods for dynamic image reconstruction from limited measurements such as video or medical imaging data.
One such important class of dynamic image reconstruction problems arises in dynamic magnetic resonance imaging (dMRI), where the data are inherently or naturally undersampled because the object is changing as the data (samples in k-space or Fourier space of the object acquired sequentially over time) is collected.
Various techniques have been proposed for reconstructing such dynamic image sequences from limited measurements \cite{lustig2,jong3,ota1}.
Such methods may achieve improved spatial or temporal resolution by using more explicit signal models compared to conventional approaches (such as k-space data sharing in dMRI, where data is pooled in time to make sets of k-space data with sufficient samples \cite{zliang16}); these methods typically achieve increased accuracy at the price of increased computation.

Recently there has been substantial interest in developing real-time dMRI reconstruction algorithms. This literature generally falls into one (or both) of the following categories \cite{dietz2018nomenclature}: methods that model the measurements as a time sequential or adaptive process, and those that generate reconstructions with minimal physical latency and are suited for real-time use in practice. Although the second of these categories is interesting and important, methods with online or time sequential properties are valuable in their own right as they often provide benefits such as adaptability to time-varying aspects of the data and reduced memory usage and computation requirements.

While some reconstruction techniques are driven by sparsity models and assume that the 
image sequence is sparse in some transform domain or dictionary \cite{lustig2}, other methods exploit low-rank or other kinds of sophisticated models \cite{zliang16, hald11, bzhao14, peder1, locallow, ota1, Can2}.
For example, L+S methods \cite{ota1,Can2} assume that the image sequence can be decomposed as the sum of low-rank and sparse (either directly sparse or sparse in some known transform domain) components that are estimated from measurements. Dictionary learning-based approaches including with low-rank models provide promising performance for dynamic image reconstruction 
\cite{lingal16,josecab2,sravbrijefraj}. Although these methods allow adaptivity to data and provide improved reconstructions, they involve batch processing and typically expensive computation and memory use. Next, we outline our contributions, particularly a new online framework that address these issues of state-of-the-art batch reconstruction methods.

 


\subsection{Contributions} \label{subsec:contrib}

This paper investigates a framework for Online Adaptive Image Reconstruction, dubbed OnAIR, exploiting learned dictionaries. 
We model spatiotemporal patches of the underlying dynamic image sequence as sparse in an (unknown) dictionary, and we propose a method to jointly and sequentially (over time) estimate the dictionary, sparse codes, and images from streaming measurements. 
Various constraints are considered for the dictionary model such as a unitary matrix, or a matrix whose atoms/columns are low-rank when reshaped into spatio-temporal matrices.
The proposed OnAIR algorithms involve simple and efficient updates and require a small, fixed amount of data to be stored in memory at a time.
Our method allows for time-varying image models that can adapt to temporal dynamics of the underlying data. Note that we use the term ``online" in this work as it is used in the signal processing literature \cite{shalev2012online}, not to imply that it is designed to process streaming measurements with sub-real-time latency. That is, our method is online in the sense that it processes measurements in a time sequential manner and can be interpreted in the empirical risk minimization framework \cite{mairal2010online}.

Numerical experiments demonstrate the effectiveness of the proposed OnAIR methods for performing video inpainting from subsampled and noisy pixels, and dynamic MRI reconstruction from very limited k-t space measurements. The experiments show that our proposed methods are able to learn dictionaries adaptively from corrupted measurements with important representational features that improve the quality of the reconstructions compared to non-adaptive schemes. Moreover, the OnAIR methods provide better image quality than batch schemes with a single adapted dictionary, and with much lower runtimes than adaptive batch (or offline) reconstructions.

While dictionary learning was studied for image reconstruction in \cite{sravbrijefraj} and earlier works, the methods involved batch processing, where the model is learned based on patches from the entire dynamic dataset concurrently. Online dictionary learning is an important and well-studied problem in the literature as well \cite{Mai, saonli1, saonli2, wensaividosat, wen2018vidosat}; however, this work is the first to study online dictionary learning-driven reconstruction from limited dynamic imaging data. Compared to previous batch reconstruction methods, the proposed OnAIR methods enjoy a number of advantages. First, the proposed methods involve modest memory use independent of the number of frames processed, while batch learning methods such as in \cite{sravbrijefraj} have a high memory requirement because they process the entire dataset and its patches synchronously, which is infeasible for large datasets. In addition, our proposed OnAIR methods use warm start initializations and sliding windows that make the method computationally efficient, requiring only a few iterations per minibatch to update the solution/model based on new data. On the other hand, batch methods such as \cite{sravbrijefraj} require many iterations over the entire dataset to ensure convergence. As a result, OnAIR enjoys significant speedups over the batch method in \cite{sravbrijefraj} as discussed in Section~\ref{sec:exp}. The OnAIR methods also allow for dictionaries that evolve over time, thereby adapting to the dynamic changes in the object. The experiments in Section~\ref{sec:exp} show that such online learned dictionaries better model the data and outperform the batch method from \cite{sravbrijefraj} in terms of image quality. Finally, the OnAIR methods offer reduced latency compared to batch methods such as \cite{sravbrijefraj} because the data can be processed and reconstructed sequentially over time without waiting until the entire dataset is measured, which is infeasible for applications involving streaming data.\footnote{We also note that \cite{sravbrijefraj} includes a model called LASSI that incorporates a global low-rank property along with the dictionary learning model. This method was shown to provide small improvements over DINO-KAT (a dictionary learning-only method) in Tables I-III of \cite{sravbrijefraj}. Extending the OnAIR framework presented here to include such global low-rank models or other priors is an interesting problem that we leave to future work.}

Short versions of this work appeared recently in \cite{moore2017online} and \cite{ravishankar2017efficient}. 
This paper extends those initial works by exploring OnAIR methods with multiple constraints on the learned models including a Unitary Dictionary (OnAIR-UD) constraint, and a Low-rank Dictionary atom (OnAIR-LD) constraint. Such constraints may offer a degree of algorithmic efficiency and robustness to artifacts in limited data settings.
For example, the updates per time instant in OnAIR-UD are simpler and non-sequential, enabling it to be faster in practice than the other OnAIR schemes.
Importantly, this paper reports extensive numerical experiments investigating and evaluating the performance of the proposed online methods in multiple inverse problems, and comparing their performance against recent related online and batch methods in the literature.
Finally, we also compare the performance of the OnAIR methods to an oracle online scheme, where the dictionary is learned offline from the ground truth data, and show that the proposed online learning from highly limited and corrupted streaming measurements performs comparably to the oracle scheme.

\subsection{Organization} \label{subsec:org}
The rest of this paper is organized as follows. Section~\ref{sec:dl} reviews the dictionary learning framework that forms the basis of our online schemes.
Section~\ref{sec:prob} describes our formulations for online adaptive image reconstruction and Section~\ref{sec:algo} presents algorithms for solving the problems.
Section~\ref{sec:exp} presents extensive numerical experiments that demonstrate the promising performance of our proposed methods in inverse problem settings such as video inpainting and dynamic MRI reconstruction from highly limited data. Finally, Section~\ref{sec:conclusion} concludes with proposals for future work.

\section{Dictionary Learning Models} \label{sec:dl}

Here, we briefly review some prior works on dictionary learning that helps build our OnAIR framework.
Given a set of signals (or vectorized image patches) that are represented as columns of a matrix $ P \in \mathbb{C}^{n \times M} $, the goal of dictionary learning (DL) is to learn a dictionary $D \in \mathbb{C}^{n \times m}$ and a matrix $Z \in \mathbb{C}^{m \times M}$ of sparse codes such that $P \approx DZ$. Traditionally, the DL problem is often formulated \cite{dl1:elad} as follows:
\begin{align} \label{eq:dl:orig}
\displaystyle\min_{D,Z} & ~~ \|P - DZ\|_F^2 \\
\text{s.t.} & ~~ \|d_i\|_2 = 1, ~ \|z_l\|_0 \leq s, ~ \forall i,l, \nonumber
\end{align}
where $d_{i}$ and $z_l$ denote the $i$th column of $D$ and the $l$th column of $Z$, respectively, and $s$ denotes a target sparsity level for each signal. Here, the $\ell_0$ ``norm" counts the number of non-zero entries of its argument, and the columns of $D$ are set to unit norm to avoid scaling ambiguity between $D$ and $Z$ \cite{dl1:kar}. Alternatives to \eqref{eq:dl:orig} exist that replace the $\ell_0$ ``norm" constraint with other sparsity-promoting constraints, or enforce additional properties on the dictionary \cite{dl1:barchi1,dl1:irami,dl1:zibul}, or enable online dictionary learning \cite{Mai}.

Dictionary learning algorithms \cite{dl1:eng,dl1:elad,dl1:Yagh,dl1:skret,Mai,dl1:smith1} typically attempt to solve \eqref{eq:dl:orig} or its variants in an alternating manner by  performing a sparse coding step (updating $Z$) followed by a dictionary update step (updating $D$). Methods such as K-SVD \cite{dl1:elad} also partially update the coefficients in $Z$ in the dictionary update step, while some recent methods update the variables jointly and iteratively \cite{dl1:directdl11}. Algorithms for \eqref{eq:dl:orig} typically repeatedly update $Z$ (involves an NP-hard sparse coding problem) and tend to be computationally expensive.



The DINO-KAT learning problem \cite{saibriarajjeff2} is an alternative dictionary learning framework that imposes a low-rank constraint on reshaped dictionary atoms (columns). The corresponding problem formulation is
\begin{align} \label{eq:dl:dinokat} 
\displaystyle\min_{D,Z} & ~~ \|P - DZ\|_F^2 + \lambda^2 \|Z\|_0 \\
\text{s.t.} & ~~ \rank(\rs(d_i)) \leq r, ~ \|d_i\|_2 = 1, ~ \|z_l\|_{\infty} \leq L, ~ \forall i, l, \nonumber
\end{align}
where the $\ell_0$ ``norm" counts the total number of nonzeros in $Z$. 
The operator $\rs(\cdot)$ reshapes dictionary atoms $d_i \in \C^n$ into matrices of size $n_1 \times n_2$ for some $n_1$ and $n_2$ such that $n = n_1 n_2$, and $r > 0$ is the maximum allowed rank for each reshaped atom. The dimensions of the reshaped atoms can be chosen on an application-specific basis. For example, when learning $D$  from 2D image patches, the reshaped atoms can have the dimensions of the 2D patch.
In the case where spatiotemporal (3D) patches are vectorized and extracted from dynamic data, the atoms could be reshaped into space-time (2D) matrices. Spatiotemporal patches of videos often have high temporal correlation, so they may be well represented by a dictionary with low-rank space-time (reshaped) atoms \cite{saibriarajjeff2}.
The parameter $\lambda > 0$ in \eqref{eq:dl:dinokat} controls the overall sparsity of $Z$. Penalizing the overall or aggregate sparsity of $Z$ enables variable sparsity levels across training signals (a more flexible model than \eqref{eq:dl:orig}). The $\ell_{\infty}$ constraints for $L > 0$ prevent pathologies (e.g., unbounded algorithm iterates) due to the non-coercive objective \cite{sairajfes}. In practice, we set $L$ very large, and the constraint is typically inactive. We proposed an efficient algorithm in \cite{saibriarajjeff2} for Problem \eqref{eq:dl:dinokat} that requires less computation than algorithms such as K-SVD.

Another alternative to the DL problems in \eqref{eq:dl:orig} and \eqref{eq:dl:dinokat} involves replacing the constraints in \eqref{eq:dl:dinokat} with the unitary constraint $D^{H}D=I$, where $I$ is the $n\times n$ identity matrix. Learned unitary operators work well in image denoising and more general reconstruction problems \cite{sabres3,sravTCI1}. Moreover, algorithms for learning unitary dictionaries tend to be computationally cheap \cite{sabres3}. Our OnAIR framework exploits some of the aforementioned dictionary structures.



\section{Problem Formulations} \label{sec:prob}



This section formulates data-driven online image reconstruction.
First, we propose an online image reconstruction framework based on an adaptive dictionary regularizer as in \eqref{eq:dl:dinokat}. Let $\{f^t \in \C^{N_x \times N_y}\}$ denote the sequence of dynamic image frames to be reconstructed. We assume that noisy, undersampled linear measurements of these frames are observed. We process the streaming measurements in minibatches, with each minibatch containing measurements of $\tilde{M} \geq 1$ consecutive frames. Let $x^t$ denote the vectorized version of the 3D tensor obtained by (temporally) concatenating $\tilde{M}$ consecutive frames of the unknown dynamic images. In practice, we construct the sequence $\{x^t\}$ using a sliding window (over time) strategy, which may involve overlapping or non-overlapping minibatches. We model the spatiotemporal (3D) patches of each $x^{t}$ as sparse with respect to a latent dictionary $D$. Under this model, we propose to solve the following online dictionary learning-driven image reconstruction problem for each time $t=1,2,3,\ldots$
\begin{align}
\nonumber (\text{P1}) ~~ & \big\{\hat{x}^t,\hat{D}^t,\hat{Z}^t\big\} =
\underset{x^t,D,Z^t}{\arg \min} ~ \frac{1}{K_t}\sum_{j=1}^t \rho^{t-j} \|y^j - A^j x^j\|_2^2 \\
\nonumber + & ~~ \frac{\lambda_S}{K_t} \displaystyle\sum_{j=1}^t \rho^{t-j} \bigg(
\displaystyle\sum_{l=1}^{M}\|P_l x^j - D z_l^j \|_2^2 + \lambda_Z^2 \|Z^j\|_0 \bigg) \\[2pt]
\nonumber \text{s.t.} & ~~ \|z_l^t\|_{\infty} \leq L, ~ \rank(\rs(d_i)) \leq r, ~ \|d_i\|_2 = 1, ~ \forall i,l.
\end{align}
Here $j$ indexes time, and $y^{t}$ denotes the (typically undersampled) measurements that are related to the underlying frames $x^t$ (that we would like to reconstruct) through the linear sensing operator $A^t$. For example, in video inpainting, $A^t$ samples a subset of pixels in $x^t$, or in dynamic MRI, $A^t$ corresponds to an undersampled Fourier encoding. Typically---and, in particular, in this work---the subsampling occurs per frame using a different random sampling pattern for each frame. For example, in dMRI, measurements are pooled in time to form sets of k-space data (the corresponding underlying object is often written in the form of a Casorati matrix \cite{zliang16}, whose rows represent voxels and  columns denote temporal frames). The operator $P_l$ is a patch extraction matrix that extracts the $l$th $n_x \times n_y \times n_t$ spatiotemporal patch from $x^t$ as a vector. A total of $M$ (possibly) overlapping 3D patches are assumed. Matrix $D \in \C^{n \times m}$ with $n = n_x n_y n_t$ is the synthesis dictionary to be learned and $z_l^t \in \C^m$ is the unknown sparse code for the $l$th patch of $x^t$, with $P_l x^t \approx D z_l^t$. Matrix $Z^t$ has $z_l^t$ as its columns, and the $\ell_{0}$ ``norm" in (P1) penalizes the aggregate sparsity (i.e., the total number of nonzeros) of $Z^{t}$. The weights $\lambda_S, \lambda_Z \geq 0$ are regularization parameters that control the relative adaptive dictionary regularization and sparsity of $Z$, respectively, in the model.
The parameter $0 < \rho \leq 1$ is an exponential forgetting factor that controls the influence of old (previous values of $t$) data in (P1), and $K_t = \sum_{j=1}^t \rho^{t-j}$ is a normalization constant for the objective.

Problem (P1) has a form often used for online optimization \cite{Mai, bot}.
The objective is a weighted average of the function $h(y^t, D, x^t, Z^t) = \|y^t - A^t x^t\|_2^2 + \lambda_S \begin{pmatrix}
\displaystyle\sum_{l=1}^{M}\|P_l x^t - D z_l^t \|_2^2 + \lambda_Z^2 \|Z^t\|_0
\end{pmatrix}$, with the constraints defining the feasible sets for $D$ and the sparse codes. 
$K_t$ serves as a normalization of the objective so that the cost can be viewed as a weighted average of the instantaneous cost $h(y^t, D, x^t, Z^t)$.\footnote{We include the normalization constant $K_t$ in (P1) to parallel the cost functions in online optimization works in the literature \cite{Mai, bot}.}\footnote{The minimization in (P1) (and later in (P2)) is only with respect to the variables $x^t$, $D$, and $Z^t$ for time $t$, so one could omit the summation terms involving $x^j$ and $Z^j$ for $j < t$ if desired. However, we retain these terms to emphasize the connection to existing methods in the online optimization literature \cite{Mai, bot}.} When $\rho=1$, this normalization also prevents the cost from becoming unbounded as $t \to \infty$. In practice, this is not an issue when processing finite-size time-series data or with $\rho<1$. Hence, for simplicity, we may omit this normalization constant.

Problem (P1) jointly estimates the adaptive dictionary model for the patches of $x^t$ together with the underlying image frames. Note that, for each time index $t$, we solve (P1) only for the latest group of frames $x^{t}$ and the latest sparse coefficients $Z^t$, while the previous images and sparse coefficients are set to their estimates from previous minibatches (\ie $x^j = \hat{x}^j$ and $Z^j = \hat{Z}^j$ for $j < t$).  However, the dictionary $D$ is adapted to all the spatiotemporal patches observed up to time $t$. We emphasize the global dependence of $D$ on all previous data by using the optimization variable $D$ within the objective rather than a time-indexed variable as done for the variables $x^t$ and $Z^t$. Assuming $\rho=1$, the objective in (P1) with respect to $D$ acts as a surrogate (upper bound) for the usual empirical (batch) risk function \cite{Mai, bot} that uses the optimal reconstructions and sparse codes (i.e., those that minimize the cost) for all the data. The exponential factor $\rho^{t-j}$ diminishes the influence of ``old'' data on the dictionary adaptation process. When the dynamic object or scene changes slowly over time, a large $\rho$ (close to 1) is preferable so that past information has more influence and vice versa.

As written in (P1), the dictionary $D$ is updated based on patches from all previous times; however, the proposed algorithm does not need to store this information during optimization. Indeed, our algorithm in Section~\ref{sec:algo} computes only a few constant-sized matrices that contain the necessary cumulative (over time) information to solve (P1).

When minibatches $x^t$ and $x^{t+1}$ do not overlap (\ie no common frames), each frame $f^t$ is reconstructed exactly once in its corresponding window in (P1). However, it is often beneficial to construct $x^t$ using an overlapping sliding window strategy \cite{wensaividosat}, in which case a frame $f^t$ may be reconstructed in multiple windows (minibatches of frames). In this case, we independently produce estimates $\hat{x}^t$ for each time index as indicated in (P1), and then we produce a final estimate of the underlying frame $f^t$ by computing a weighted average of the reconstructions of that frame from each window in which it appeared. We found empirically that an exponentially $\rho$-weighted average (similar to that in (P1)) performed better than alternatives such as an unweighted average of the estimates from each window or using only the most recent reconstruction from the latest window.

In (P1), we imposed a low-rank constraint on the dictionary atoms. As an alternative, we consider constraining the dictionary to be a unitary matrix. 
The online optimization problem in this case is as follows:
\begin{align} 
\nonumber (\text{P2}) ~~ & \big\{\hat{x}^t,\hat{D}^t,\hat{Z}^t\big\} =
\underset{x^t,D,Z^t}{\arg \min} ~ \frac{1}{K_t}\sum_{j=1}^t \rho^{t-j} \|y^j - A^j x^j\|_2^2 \\
\nonumber & \hspace{-0.05in} ~~ + \frac{\lambda_S}{K_t} \displaystyle\sum_{j=1}^t \rho^{t-j} \bigg(
\displaystyle\sum_{l=1}^{M}\|P_l x^j - D z_l^j \|_2^2 + \lambda_Z^2 \|Z^j\|_0 \bigg) \\[2pt]
\nonumber \text{s.t.} & ~~ D^H D = I,
\end{align}
where all terms in (P2) are defined as in (P1). Note that (P2) does not require the $\ell_{\infty}$-norm constraints on the sparse coefficients $Z^t$ because the unitary constraint on the dictionary precludes the possibility of repeated dictionary atoms, which was the motivation for including these constraints in (P1) \cite{sairajfes}.

\section{Algorithms and Properties} \label{sec:algo}

\begin{figure}[t]
\begin{center}
\includegraphics[width=\linewidth]{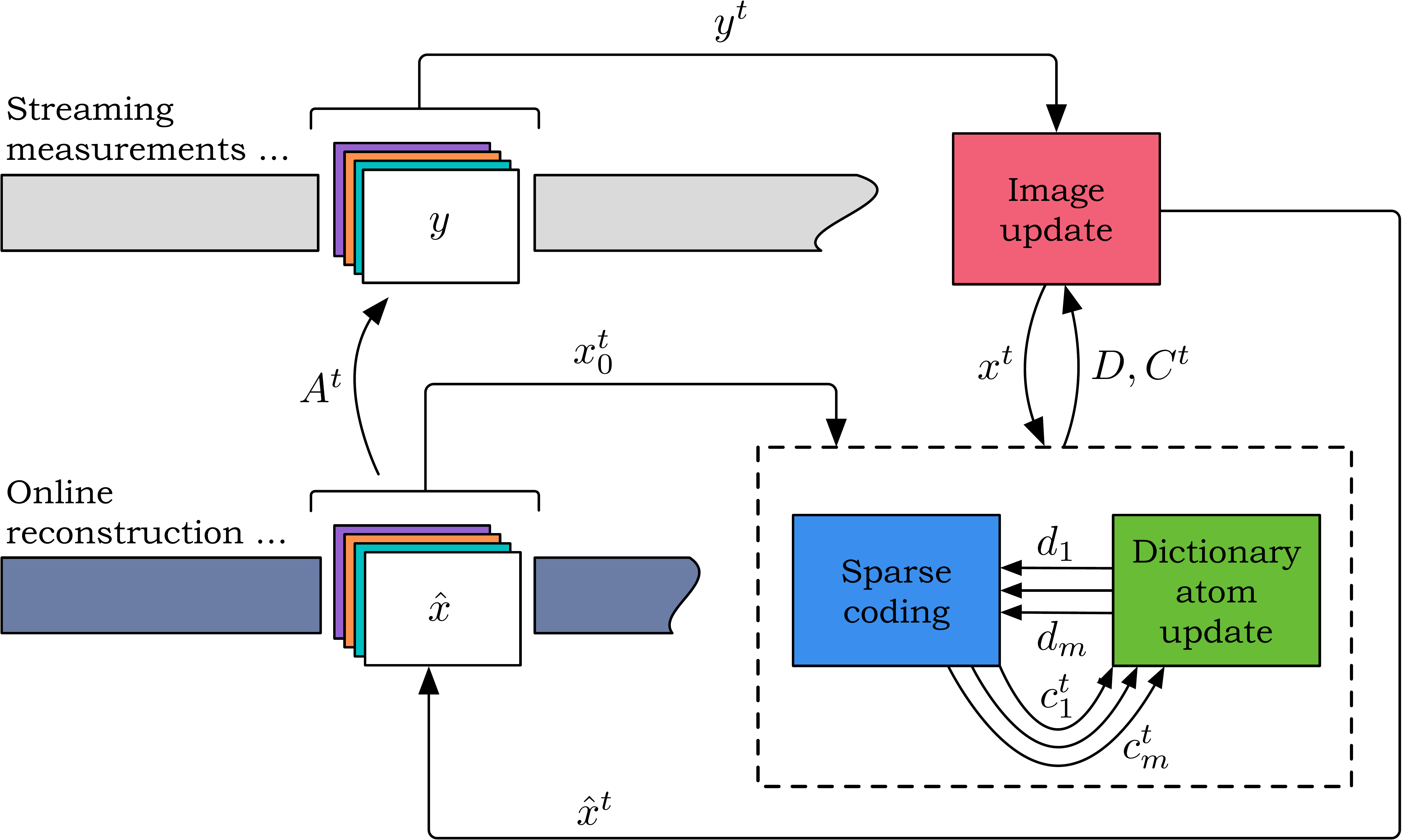}
\caption{Flowchart of the proposed online adaptive dictionary learning-driven image reconstruction scheme at time $t$. The input is a vector $y^t$ containing the streaming measurements for the current minibatch of frames, $\hat{x}^t$ denotes the corresponding reconstructed minibatch, and $x^{t}$ denotes intermediate reconstructions in the alternating scheme. In the \emph{dictionary learning step} $(D, C^t)$ are updated with $x^t$ held fixed, by performing block coordinate descent over the columns of $C^t$ (\emph{sparse coding}) and the columns of $D$ (\emph{dictionary atom update}). Then, the frames $x^t$ are updated (the \emph{image update step}) with $(D, C^t)$ held fixed. This process is repeated a few times, and the final frame estimates $\hat{x}^t$ are integrated into the streaming reconstruction, $\hat{x}$. $x_{0}^{t}$ denotes the initialization (warm start).}
\label{fig:flowchart}
\end{center}
\end{figure}

This section presents the algorithms for Problems (P1) and (P2) and their properties.
We propose an alternating minimization-type scheme for (P1) and (P2) and exploit the online nature of the optimization to minimize the costs efficiently. 
At each time index $t$, we alternate a few times between updating $(D, Z^t)$ while holding $x^t$ fixed (the \emph{dictionary learning step}) and then updating $x^t$ with $(D, Z^t)$ held fixed (the \emph{image update step}). 
For each $t$, we use a warm start for (initializing) the alternating approach.
We initialize the dictionary $D$ with the most recent dictionary ($\hat{D}^{t-1}$). Frames of $x^t$ that were estimated in the previous (temporal) windows are initialized with the most recent $\rho$-weighted reconstructions, and new frames are initialized using simple approaches (\eg interpolation in the case of inpainting). Initializing the sparse coefficients $Z^t$ with the codes estimated in the preceding window ($\hat{Z}^{t-1}$) worked well in practice. All updates are performed efficiently and with modest memory usage as will be shown next. Fig.~\ref{fig:flowchart} provides a graphical flowchart depicting our proposed online scheme. The next subsections present the alternating scheme for each time $t$ in more detail.


\subsection{Dictionary Learning Step for (P1)} \label{subsec:dstep}
Let $C^t \triangleq (Z^t)^{H}$. Minimizing (P1) with respect to $(D, C^t)$ yields the following optimization problem:
\begin{align}
\displaystyle\min_{D,C^t} & ~~ \displaystyle\sum_{j=1}^t \rho^{t-j} \|P^j - D(C^j)^H \|_F^2 + \lambda_Z^2 \|
C^t\|_0 \label{eq:dstep} \\
\text{s.t.} & ~~ \|c_i^t\|_{\infty} \leq L, ~\rank(\rs(d_i)) \leq r, ~ \|d_i\|_2=1 ~ \forall i, \nonumber
\end{align}
where $P^{j} \in \C^{n \times M}$ is the matrix whose columns contain the patches $P_l x^j$ for $1 \leq l \leq M$, and $c_i^t$ is the $i$th column of $C^t$. We use a block coordinate descent approach \cite{sairajfes} (with few iterations) to update the sparse coefficients $c_i^t$ and atoms $d_i$ (columns of $D$) in \eqref{eq:dstep} sequentially. For each $1 \leq i \leq m$, we first minimize \eqref{eq:dstep} with respect to $c_i^t$ keeping the other variables fixed (\emph{sparse coding step}), and then we update $d_i$ keeping the other variables fixed (\emph{dictionary atom update step}). These updates are performed in an efficient online manner as described in the following subsections.

Note that to theoretically converge to a solution of Problem \eqref{eq:dstep}, the block coordinate descent updates described here would have to be iterated many times. Fortunately, exhaustive subproblem updates are not necessary in practice for online sliding window optimization problems \cite{Mai, bot}, because warm starts of variables from the previous window(s) can be used as high quality initializations for the updates in the current time window. As such, a few iterations suffice to further update the variables based on new data. In the proposed OnAIR algorithm, the dictionary is continuously updated over many iterations to adapt to the time-series of measurements. If the streaming data all agree with a common dictionary model, it is reasonable to expect that the online dictionary updates presented below will converge over time.

\subsubsection{Sparse Coding Step} \label{subsubsec:cstep}
Minimizing \eqref{eq:dstep} with respect to $c_i^t$ leads to the following subproblem:
\begin{equation} \label{eqop5}
\begin{array}{r@{~}l}
\displaystyle\min_{c_i^t \in \C^M} & ~~ \|E_i^t - d_i(c_i^t)^H\|_F^2 + \lambda_Z^2 \|c_i^t\|_0 \\
\text{s.t.} & ~~ \|c_i^t\|_{\infty} \leq L,
\end{array}
\end{equation}
where the matrix
\begin{equation} \label{eq:Ei:matrix}
E_i^t := P^t - \sum_{k\neq i} d_k(c_k^t)^H
\end{equation}
is defined based on the most recent estimates of the other atoms and sparse coefficients. The solution to \eqref{eqop5}, assuming $L > \lambda_Z$, is given by \cite{sairajfes}
\begin{equation} \label{tru1ch4}
\hat{c}_{i}^{t} =  \min\left(|H_{\lambda_Z}((E_{i}^{t})^{H}d_{i})|, \ L 1_{M}\right) \odot e^{j \angle (E_i^t)^H d_i},
\end{equation}
where $H_{\lambda_Z}(\cdot)$ is the elementwise hard thresholding operator that sets entries with (complex) magnitude less than $\lambda_Z$ to zero and leaves other entries unaffected, $1_M$ is a length-$M$ vector of ones, $\odot$ and $\min(\cdot, \cdot)$ denote elementwise multiplication and elementwise minimum respectively, and $e^{j \angle \cdot}$ is computed elementwise, with $\angle$ denoting the phase. We do not construct $E_i^t$ in \eqref{tru1ch4} explicitly; rather we efficiently compute the matrix-vector product $(E_i^t)^H d_i = (P^t)^H d_i - C^t D^H d_i + c_i^t$ based on the most recent estimates of each quantity using sparse matrix-vector operations \cite{sairajfes}.

\subsubsection{Dictionary Atom Update Step} \label{subsubsec:dstep}
Here, we minimize \eqref{eq:dstep} with respect to $d_i$. This update uses past information via the forgetting factor $\rho$. Let $\tilde{P}^j := \sqrt{\rho^{t-j}} P^j$ and $\tilde{C}^j := \sqrt{\rho^{t-j}} C^j$ denote the $\rho$-weighted patches and sparse coefficients, respectively, and let $\tilde{P}^{1:t}$ and $\tilde{C}^{1:t}$ denote the matrices formed by stacking the $\tilde{P}^j$'s horizontally and $\tilde{C}^j$'s vertically, respectively, for times 1 to $t$.
Finally, define $\tilde{E}_i^{1:t} := \tilde{P}^{1:t} - \sum_{k \neq i} d_k (\tilde{c}_k^{1:t})^H$
using the most recent estimates of all variables, with $\tilde{c}_{k}^{1:t}$ denoting the $k$th column of $\tilde{C}^{1:t}$. Using this notation, the minimization of \eqref{eq:dstep} with respect to $d_i$ becomes
\begin{equation} \label{eqop6}
\begin{array}{r@{~}l}
\displaystyle\min_{d_i \in \C^n} ~ & \|\tilde{E}_i^{1:t} - d_i (\tilde{c}_i^{1:t})^H\|_F^2 \\[6pt]
\text{s.t.} ~ & \rank(\rs(d_i)) \leq r, ~ \|d_i\|_2 = 1.
\end{array}
\end{equation}
Let $U_r \Sigma_r V_r^H$ be the rank-$r$ truncated singular value decomposition (SVD) of the matrix $\rs(\tilde{E}_i^{1:t}\tilde{c}_i^{1:t})$ that is obtained by computing the $r$ leading singular vectors and singular values of the full SVD $\rs(\tilde{E}_i^{1:t}\tilde{c}_i^{1:t}) := U \Sigma V^H$. Then a solution to \eqref{eqop6} is given by \cite{saibriarajjeff2}
\begin{equation} \label{tru1ch4g}
\rs(\hat{d}_i) = \begin{cases}
\dfrac{U_r \Sigma_r V_r^H}{\|\Sigma_r\|_F}, & \text{if } \tilde{c}_i^{1:t} \neq 0 \\[8pt]
\quad ~ W , & \text{if } \tilde{c}_i^{1:t} = 0,
\end{cases}
\end{equation}
where $W$ is any matrix of appropriate dimension with rank at most $r$ such that $\|W\|_F = 1$. In our experiments, we set $W$ to be the reshaped first column of the $n \times n$ identity matrix, which worked well.

The main computation in \eqref{tru1ch4g} is computing $\tilde{E}_i^{1:t}\tilde{c}_i^{1:t}$, since the SVD of the small $n_y n_x \times n_t$ (i.e., space-time matrix with $n_t < n_y n_x$ typically) matrix $\rs(\tilde{E}_i^{1:t}\tilde{c}_i^{1:t})$ has negligible computational cost. In principle, the matrix-vector multiplication $\tilde{E}_i^{1:t}\tilde{c}_i^{1:t}$ depends on all past information processed by the streaming algorithm; however, it can be recursively computed using constant time and memory. Indeed, observe that
\begin{align} \label{eq:ec:recurse}
\tilde{E}_{i}^{1:t}\tilde{c}_{i}^{1:t} =&
\displaystyle\sum_{j=1}^t \tilde{E}_i^j \tilde{c}_i^j = \displaystyle\sum_{j=1}^t \rho^{t-j} E_i^j c_i^j \\
=& \displaystyle\sum_{j=1}^t \rho^{t-j} \left(P^j - D(C^j)^H + d_i (c_i^j)^H\right)c_i^j \nonumber \\
=& \underbrace{\Bigg[ \displaystyle\sum_{j=1}^t \rho^{t-j} P^j c_i^j \Bigg]}_{=: q_i^t} -
D \underbrace{\Bigg[ \displaystyle\sum_{j=1}^t \rho^{t-j} (C^j)^H c_i^j \Bigg]}_{=: g_i^t} + \nonumber \\
& \hspace{1.3cm} d_i \underbrace{\Bigg[ \displaystyle\sum_{j=1}^t \rho^{t-j} \|c_i^j\|^2 \Bigg]}_{=[g_i^t]_i}, \nonumber
\end{align}
where $[z]_i$ denotes the $i$th element of a vector $z$. The vectors $q_i^t$ and $g_i^t$
depend on all previous data, but they can be recursively computed over time as
\begin{equation} \label{eq:fg:recurse}
\begin{array}{r@{~}l}
q_i^t =& \rho q_i^{t-1} + P^t c_i^t, \\[3pt]
g_i^t =& \rho g_i^{t-1} + (C^t)^H c_i^t,
\end{array}
\end{equation}
where for each column index $i$, the matrix $C^t$ is understood to contain the latest versions of the sparse codes already updated (sequentially) during the dictionary learning step. Using these recursive formulae, the product $\tilde{E}_i^{1:t}\tilde{c}_i^{1:t}$ can be readily computed each time in our algorithm. Thus, the update in \eqref{tru1ch4g} can be performed in a fully online manner (i.e., without storing all the past data).

Alternatively, we can collect the vectors $q_i^{t-1}$ and $g_i^{t-1}$ as columns of matrices $Q^{t-1} \in \C^{n \times m}$ and $G^{t-1} \in \C^{m \times m}$, and perform the following recursive updates once at the end of 
of the overall algorithm for (P1) (i.e., once per minibatch of frames):
\begin{equation} \label{eq:FG:recurse}
\begin{array}{r@{~}l}
Q^t =& \rho Q^{t-1} + P^t C^t \\[3pt]
G^t =& \rho G^{t-1} + (C^t)^H C^t.
\end{array}
\end{equation}
Here $C^t$ denotes the final sparse codes estimate 
from the algorithm for (P1). 
In this case, when performing the inner update \eqref{eq:ec:recurse}, the contributions of the two terms on the right hand side of \eqref{eq:fg:recurse} are incorporated separately.
The matrices $Q^t \in \C^{n \times m}$ and $G^t \in \C^{m \times m}$ are small, constant-sized matrices whose dimensions are independent of the time index $t$ and the dimensions of the frame sequence, so they are stored for efficient use in the next minibatch. Moreover, the matrix $C^t$ is sparse, so all the matrix-matrix multiplications in \eqref{eq:FG:recurse} (or the matrix-vector multiplications in \eqref{eq:fg:recurse}) are computed using efficient sparse operations.

\subsection{The Unitary Dictionary Variation} \label{subsec:unitary}


In the case of (P2), unlike for (P1), we do not perform block coordinate descent over the columns of $D$ and $(Z^t)^H$. Instead we minimize (P2) with respect to each of the matrices $Z^t$ and $D$ directly and exploit simple closed-form solutions for the matrix-valued updates. The following subsections describe the solutions to the $Z^t$ and $D$ update subproblems.

\subsubsection{Sparse Coding Step}
Since $D$ is a unitary matrix, minimizing (P2) with respect to $Z^t$ yields the following subproblem:
\begin{equation} \label{eq:unitary:z}
\displaystyle\min_{Z^t \in \mathbb{C}^{n \times M}} ~ \|D^H P^t - Z^t\|_F^2 + \lambda_Z^2 \|Z\|_0,
\end{equation}
where $P^t$ is again the matrix whose $l$th column is $P_l x^t$. 
The solution to \eqref{eq:unitary:z} is given by the elementwise hard-thresholding (at threshold $\lambda_Z$) operation
\begin{equation} \label{eq:unitary:zsol}
\hat{Z}^{t} = H_{\lambda_Z}(D^H P^t).
\end{equation}

\subsubsection{Dictionary Update Step}
Minimizing (P2) with respect to $D$ results in the following optimization:
\begin{equation} \label{eq:unitary:d}
\begin{array}{r@{~}l}
\displaystyle\min_{D \in \mathbb{C}^{n \times n}} ~ & \displaystyle\sum_{j=1}^t \rho^{t-j} \|P^j - D (C^j)^H\|_F^2 \\[14pt]
\text{s.t.} ~ & D^H D = I,
\end{array}
\end{equation}
where $C^j \triangleq (Z^j)^H$ is used for notational convenience. Using the definitions of the matrices $\tilde{P}^{1:t}$ and $\tilde{C}^{1:t}$ from the dictionary atom updates in Section~\ref{subsec:dstep}, we can equivalently write \eqref{eq:unitary:d} as
\begin{equation} \label{eq:unitary:d2}
\displaystyle\min_{D \in \mathbb{C}^{n \times n}}  \|\tilde{P}^{1:t} - D (\tilde{C}^{1:t})^H\|_F^2 \;\; \text{s.t.} \;\; D^{H} D = I.
\end{equation}
Problem~\eqref{eq:unitary:d2} is a well-known orthogonal Procrustes problem \cite{gower1975generalized}. The solution is given by 
\begin{equation}
\hat{D} = UV^{H}, \label{dunitaryupdate} 
\end{equation}
where $U \Sigma V^{H}$ is a full SVD of $\tilde{P}^{1:t} \tilde{C}^{1:t}$.
Similarly as in \eqref{eq:FG:recurse}, let $Q^t = \tilde{P}^{1:t} \tilde{C}^{1:t}$.
The matrix $Q^t \in \C^{n \times n}$ can be recursively updated over time according to \eqref{eq:FG:recurse} (or \eqref{eq:fg:recurse}), so the dictionary update step can be performed efficiently and fully online.

\subsection{Image Update Step} \label{subsec:xstep}
Minimizing (P1) or (P2) with respect to the minibatch $x^{t}$ yields the following quadratic sub-problem:
\begin{equation} \label{eq:xstep}
\min_{x^t \in \mathbb{C}^{N_{x} N_{y} \tilde{M}}} ~ \|A^t x^t - y^t\|_2^2 + \lambda_S \displaystyle\sum_{l=1}^M \| P_l x^t - D z_l^t \|_2^2.
\end{equation}
Problem \eqref{eq:xstep} is a least squares problem with normal equation
\begin{equation} \label{rowq2}
\bigg((A^t)^H A^t + \lambda_S \sum_{l=1}^M P_l^{T} P_l \bigg) x^t = (A^t)^H y^t + \lambda_S \sum_{l=1}^M P_l^{T} D z_l^t.
\end{equation}
In applications such as video denoising or inpainting, the matrix pre-multiplying $x^t$ in \eqref{rowq2} is a
diagonal matrix that can be efficiently pre-computed and inverted. More generally, in inverse problems where the matrix pre-multiplying $x^t$ (i.e., $(A^t)^H A^t$ in particular) in \eqref{rowq2} is not diagonal or readily diagonalizable (\eg in dynamic MRI with multiple coils), we minimize \eqref{eq:xstep} by applying an iterative optimization method. One can use any classical algorithm in this case, such as the conjugate gradient (CG) method.
For the experiments shown in Section~\ref{sec:exp}, we used a few iterations (indexed by $i$) of the simple proximal gradient method \cite{pat2,parikh2014proximal} with updates of the form
\begin{equation} \label{rowq1}
x^{t,i+1} = \prox_{\tau_i \theta}\left(x^{t,i} - \tau_i (A^t)^H (A^t x^{t,i} - y^t)\right),
\end{equation}
where $\theta(x) := \lambda_S \sum_{l=1}^M \|P_l x - D z_l^t\|_2^2$ and the proximal operator of a function $h$ is 
$\prox_{h}(x) := \displaystyle\arg \min_{z} ~ 0.5 \|x-z\|_2^2 + h(z)$.
The proximal operation in \eqref{rowq1} corresponds to a simple least squares problem 
that is solved efficiently by inverting a diagonal matrix $I + 2 \tau_i \lambda_S \sum_{l=1}^M P_l^{T} P_l$ (arising from the normal equation of the proximal operation), 
which is pre-computed.
A constant step-size $\tau_i = \tau < 2/\|A^t\|_2^2$ suffices for convergence \cite{pat2}. Moreover, the iterations \eqref{rowq1} monotonically decrease the objective in \eqref{eq:xstep} when a constant step size $\tau \leq 1 / \|A^t\|_2^2$ is used \cite{parikh2014proximal}.



Fig.~\ref{im6p} summarizes the overall OnAIR algorithms for the online optimization Problems (P1) and (P2). In practice, we typically use more iterations $K$ for reconstructing the first minibatch of frames, to create a good warm start for further efficient online optimization. The initial $\hat{D}^{0}$ in the algorithm can be set to an analytical dictionary (e.g., based on the DCT or wavelets) and we set the initial $\hat{Z}^{0}$ to a zero matrix.


\begin{figure}
\begin{tabular}{p{8.3cm}}
\hline
OnAIR Algorithms for (P1) and (P2)\\
\hline
 \textbf{Inputs\;:} \:\:\: measurement sequence $\{y^t\}$, weights $\lambda_{S}$ and $\lambda_{Z}$, rank $r$, upper bound $L$, forgetting factor $\rho$, number of dictionary learning iterations $\hat{K}$, number of image update iterations $\tilde{K}$, and the number of outer iterations per minibatch $K$.\\
 \textbf{Outputs\;:} \:\: reconstructed dynamic image sequence  $\{\hat{f}^t\}$. \\
\textbf{Initialization:} \; Initial (first) estimates $\{f_{0}^t\}$ of new incoming (measured) frames, initial $\hat{D}^{0}$ and $\hat{Z}^{0}$ at $t=0$, $Q^{0} = 0^{n \times m}$ and $G^{0} = 0^{n \times m}$.\\ \vspace{-0.1in}
\textbf{For \;$t$ = $1,2,3,\ldots$, do}\\
\vspace{-0.1in}
\hspace{0.02in}\textbf{Warm Start:} \; Set $C^{t}_{0} = (\hat{Z}^{t-1})^{H}$, $D^{t}_{0}=\hat{D}^{t-1}$, and $x^{t}_{0}$ is set as follows: frames estimated in previous minibatches are set to a $\rho$-weighted average of those estimates and new frames are set as in $\{f_{0}^t\}$. \\
\hspace{0.07in}\textbf{For \;$k$ = $1:$ $K$ repeat}\\
\begin{enumerate} \vspace{-0.06in}
\item \textbf{Dictionary Learning Step:}
For (P1), update $\left ( D^{t}_{k}, C^{t}_{k} \right )$ by sequentially updating the $1\leq i \leq m$ coefficient and dictionary columns using \eqref{tru1ch4} and \eqref{tru1ch4g} (that uses \eqref{eq:ec:recurse} and \eqref{eq:fg:recurse}) respectively, with $\hat{K}$ iterations and initial $\left ( D^{t}_{k-1}, C^{t}_{k-1} \right )$.
For (P2), update $\left ( D^{t}_{k}, C^{t}_{k} \right )$ using $\hat{K}$ alternations between performing \eqref{eq:unitary:zsol} and \eqref{dunitaryupdate} with  initial $\left ( D^{t}_{k-1}, C^{t}_{k-1} \right )$.
\item \textbf{Image Update Step:} Update $x_{k}^{t}$ using \eqref{rowq2} or with $\tilde{K}$ iterations of an optimization scheme (e.g., CG or \eqref{rowq1}), with initialization $x^{t}_{k-1}$.
\end{enumerate}\\  \vspace{-0.2in}
\hspace{0.07in}\textbf{End} \\ \vspace{-0.09in}
\hspace{0.02in}\textbf{Recursive Updates:}  \; Update $Q^{t}$ and $G^{t}$ using \eqref{eq:FG:recurse}.\\
\hspace{0.02in}\textbf{Output Updates:}  \; Set $\hat{Z}^{t} =  (C^{t}_{K})^{H}$, $\hat{D}^{t}= D^{t}_{K}$ and $\hat{x}^{t} = x^{t}_{K}$. For frames not occurring in future windows, $\hat{f}^t$  is set to a $\rho$-weighted average of estimates over minibatches.\\
 \vspace{-0.06in}
\textbf{End} \\
\hline
\end{tabular}
\caption{The OnAIR algorithms for Problems (P1) and (P2), respectively. Superscript $t$ denotes the time or minibatch index.} \label{im6p}
\end{figure}






\subsection{  Cost and Convergence} \label{subsec:computation}

The computational cost for each time index $t$ of the proposed algorithms for solving the online image reconstruction problems (P1) and (P2) scales as $O(n^{2} M)$, where $D \in \mathbb{C}^{n \times m}$ with $m \propto n$ assumed, and $M$ is the number of (overlapping) patches in each temporal window. The cost is dominated by various matrix-vector multiplications. Assuming each window's length $\tilde{M} \ll n$, the memory (storage) requirement for the proposed algorithm scales as $O(n M)$, which is the space required to store the image patches of $x^t$ when performing the updates for (P1) or (P2). Typically the minibatch size $\tilde{M}$  is small (to allow better tracking of temporal dynamics), so the number and maximum temporal width of 3D patches in each window are also small, ensuring modest memory usage for the proposed online methods.


While the overall computational cost and memory requirements for each time index $t$ are similar for the algorithms for (P1) and (P2), the simple matrix-valued forms of the dictionary and sparse code updates for (P2)
result in practice, in a several-fold decrease in actual runtimes due to optimizations inherent to matrix-valued computations in modern linear algebra libraries.

We emphasize that the OnAIR framework is \textit{computationally cheap} in the sense that it requires only a small number of efficient updates in order to achieve high quality reconstructions in practice. Our proposed algorithm uses warm start initializations and sliding windows, and, at each step, we perform only a few iterations of descent updates to update the variables based on the new minibatch of data. Since the dictionary is updated continuously over time and the sparse coefficients and frames are also estimated over multiple windows, the warm start procedures produce high quality initializations that enable limited iterations in each time window. Limiting to a few iterations can also prevent over-fitting of the model to the (noise or artifacts in the) new data, which can be a concern especially for smaller values of $\rho$ (less memory). Such algorithmic strategies are also adopted in the online optimization literature \cite{Mai, bot}, where solving each instantaneous problem exhaustively is not the goal (indeed this would be infeasible when processing streaming data); instead, online algorithms efficiently improve/adapt estimates over time, usually with asymptotic convergence guarantees wherein the model converges as $t \to \infty$ to stationary points of the expectation of the instantaneous cost.

The proposed algorithms involve either exact block coordinate descent updates or for example, proximal gradient iterations (with appropriate step size) when the matrix pre-multiplying $x^{t}$ in \eqref{rowq2} is not readily diagonalizable. These updates are guaranteed to monotonically decrease the objectives in (P1) and (P2) for each time index $t$.
Whether the overall iterate sequence produced by the algorithms also converges over time (see e.g., \cite{Mai}) is an interesting open question that we leave for future work.




\begin{table*}[!t]
\resizebox{\textwidth}{!}{
\centering
\begin{tabular}{|c||c|c|c|c|c||c|c|c|c|c||c|c|c|c|c|}
\hline
\multirow{2}{*}{\% Missing Pixels} & \multicolumn{5}{c||}{Coastguard} & \multicolumn{5}{c||}{Bus} & \multicolumn{5}{c|}{Flower Garden} \\
\cline{2-16}
                   & 50\% & 60\% & 70\% & 80\% & 90\% & 50\% & 60\% & 70\% & 80\% & 90\% & 50\% & 60\% & 70\% & 80\% & 90\% \\
\hline \hline
OnAIR-FD     & 33.1 & \textbf{31.4} & \textbf{29.6} & \textbf{27.3} & 22.5 & 28.7 & 27.1 & \textbf{25.5} & \textbf{23.7} & \textbf{21.5} & \textbf{24.4} & \textbf{22.8} & \textbf{21.0} & \textbf{18.8} & 15.8 \\ \hline 
OnAIR-UD   & \textbf{33.8} & 31.3 & 28.1 & 24.8 & 21.9 & \textbf{29.7} & \textbf{27.6} & \textbf{25.5} & 23.4 & 21.1 & \textbf{24.4} & 22.1 & 19.6 & 17.1 & 15.6 \\ \hline 
Online (DCT)       & 32.7 & 30.3 & 27.8 & 25.3 & 22.6 & 28.4 & 26.7 & 25.0 & 23.1 & 20.8 & 23.3 & 21.6 & 19.9 & 18.1 & 16.3 \\ \hline 
Batch Learning     & 33.1 & 31.2 & 29.1 & 26.3 & 22.8 & 27.8 & 26.3 & 24.7 & 22.9 & 20.9 & 23.5 & 21.8 & 20.1 & 18.2 & 16.1 \\ \hline 
Interpolation (3D) & 29.8 & 28.5 & 27.3 & 25.9 & \textbf{24.1} & 27.3 & 25.7 & 24.0 & 22.1 & 20.0 & 20.6 & 19.6 & 18.5 & 17.5 & \textbf{16.4} \\ \hline
Interpolation (2D) & 28.2 & 26.5 & 24.9 & 23.1 & 21.1 & 26.0 & 24.8 & 23.7 & 22.5 & 21.1 & 20.1 & 18.8 & 17.5 & 16.2 & 14.8 \\ \hline
\end{tabular}
}
\caption{PSNR values in decibels (dB) for video inpainting for three videos from the BM4D dataset at various percentages of missing pixels. The methods considered are the proposed OnAIR-FD method (i.e., $r=5$), the proposed OnAIR-UD method, online inpainting with a fixed DCT dictionary, the batch dictionary learning-based reconstruction method with $r=5$, 2D (frame-by-frame cubic) interpolation, and 3D interpolation. The best PSNR value for each undersampling level for each video is in bold.}
\label{tab:inpaint}
\end{table*}

\section{Numerical Experiments} \label{sec:exp}
This section presents extensive numerical experiments illustrating the usefulness of the proposed OnAIR methods. We consider two inverse problem applications in this work: video reconstruction (inpainting) from noisy and subsampled pixels, and dynamic MRI reconstruction from highly undersampled k-t space data. The algorithm for (P1) is dubbed OnAIR-LD when the parameter $r$ is lower than its maximum setting (i.e., low-rank dictionary atoms) and is dubbed OnAIR-FD (i.e., full-rank dictionary atoms) otherwise. The algorithm for (P2) is dubbed OnAIR-UD (unitary dictionary). The following subsections discuss the applications and results.\footnote{The software to reproduce our results is available at \url{http://web.eecs.umich.edu/~fessler}.} See the supplementary material of this manuscript for additional numerical experiments.

\subsection{Video Inpainting} \label{subsec:inpainting}


\subsubsection{Framework}
First, we consider video inpainting or reconstruction from subsampled and potentially noisy pixels. We work with the publicly available videos\footnote{The data is available at \url{http://www.cs.tut.fi/~foi/GCF-BM3D}.} provided by the authors of the BM4D method \cite{maggioni2013nonlocal}. We process the first $150$ frames of each video at native resolution. We measure a (uniform) random subset of the pixels in each frame of the videos, and also simulate additive (zero mean) Gaussian noise for the measured pixels in some experiments. The proposed OnAIR methods are then used to reconstruct the videos from their corrupted (noisy and/or subsampled) measurements.

The parameters for the proposed OnAIR methods are chosen as follows.
We used a sliding (temporal) window of length $\tilde{M}=5$ frames with a temporal stride of 1 frame, to reconstruct (maximally overlapping) minibatches of frames. In each window, we extracted $8 \times 8 \times 5$ overlapping spatiotemporal patches with a spatial stride of 2 pixels. We learned a square $320 \times 320$ dictionary, and the operator $\rs(\cdot)$ reshaped dictionary atoms into $64 \times 5$ space-time matrices. We ran the OnAIR algorithms for (P1) and (P2)  for $K=7$ iterations in each temporal window (minibatch), with $\hat{K} = 1$ inner iteration (of block coordinate descent or alternation) for updating $(D, Z^{t})$, and used a direct reconstruction as per \eqref{rowq2} in the image update step. A forgetting factor $\rho = 0.9$ was observed to work well, and we ran the algorithms for more ($K=50$) iterations for the first minibatch of data. We initialized dictionary $D$ as the transpose of the 3D discrete cosine transform (DCT) matrix\footnote{The 3D DCT matrix used here is obtained as the Kronecker product of three one-dimensional DCTs of dimensions matching the patch dimensions ($8 \times 8 \times 5$). Applying the (separable) 3D DCT transform to a patch corresponds to applying 1D DCTs along the row, column, and temporal dimensions of the patch. Equivalently, the patch is sparse coded in the dictionary $D$ that is the transpose of the 3D DCT.} (a sparsifying transform), and the initial sparse codes were zero.

Newly arrived frames were first initialized (i.e., in the first minibatch they occur) using 2D cubic interpolation. We simulated various levels of subsampling of the videos (with and without noise), and we chose $r=5$ or full-rank atoms in (P1), which outperformed low-rank atoms in the experiments here.
The videos in this section have substantial temporal motion, so allowing full-rank atoms enabled the algorithm to better learn the dynamic temporal features of the data. Section~\ref{subsec:dmri} demonstrates the usefulness of low-rank atoms in (P1).
We tuned the weights $\lambda_{S}$ and $\lambda_{Z}$ for (P1) and (P2) to achieve good reconstruction quality at an intermediate undersampling factor (70\% missing pixels) for each video, and used the same parameters for other factors.



We measure the performance of our methods using the (3D) peak signal-to-noise ratio (PSNR) metric that is computed as the ratio of the peak pixel intensity in the video to the root mean square reconstruction error relative to the ground truth video. All PSNR values are expressed in decibels (dB).

We compare the performance of the proposed OnAIR-FD ($r=5$) and OnAIR-UD methods for (P1) and (P2) with that of an identical online algorithm but which uses a fixed DCT dictionary. We refer to this variation henceforward as the Online DCT method. In our experiments, we implemented the Online DCT method as outlined in Fig.~\ref{im6p}, but omitting the dictionary atom update step in \eqref{tru1ch4g}. We also produce reconstructions using a ``batch" version of the method for (P1) \cite{saibriarajjeff2} (with $r=5$) that processes all video frames jointly; this method is equivalent to the proposed online method for (P1) with $\tilde{M}$ set to the total number of frames in the video. For each comparison method, we used the same patch dimensions, initializations, etc., and tuned the parameters $\lambda_{S}$ and $\lambda_{Z}$ of each method individually. 
The one exception is that we used a spatial patch stride of 4 pixels for the Batch Learning method rather than the 2 pixel spatial stride used for the online methods. This change was made because the batch method is memory intensive---it extracts and processes image patches from the entire video concurrently---so it was necessary to process fewer patches to make the computation feasible. We ran the batch learning-based reconstruction for 20 iterations, which worked well. 
Finally, we also compare the OnAIR reconstructions to baseline reconstructions produced by 2D (frame-by-frame cubic) interpolation and 3D interpolation (using the natural neighbor method in MATLAB).

\begin{table}[t!]
\centering
\resizebox{0.9\linewidth}{!}{
\begin{tabular}{|c||c|c|c|c|c|}
\hline
\multirow{2}{*}{\% Missing Pixels} & \multicolumn{5}{c|}{Coastguard (25 dB PSNR)} \\
\cline{2-6}
                   & 50\% & 60\% & 70\% & 80\% & 90\% \\
\hline \hline
OnAIR-FD    & 28.6 & 27.9 & \textbf{27.2} & \textbf{26.1} & \textbf{23.9} \\ \hline 
OnAIR-UD   & \textbf{29.4} & \textbf{28.6} & 26.6 & 24.9 & 22.0 \\ \hline 
Online (DCT)       & 28.6 & 27.8 & 26.7 & 25.1 & 22.9 \\ \hline 
Batch Learning     & 28.6 & 27.8 & 26.6 & 25.1 & 22.5 \\ \hline 
Interpolation (3D) & 26.2 & 25.9 & 25.4 & 24.6 & 23.4 \\ \hline
Interpolation (2D) & 24.7 & 24.0 & 23.0 & 21.9 & 20.2 \\ \hline
\end{tabular}
}
\caption{PSNR values in decibels (dB) for video inpainting for the Coastguard video corrupted by Gaussian noise with 25dB PSNR, at various percentages of missing pixels. The following methods are compared: the proposed OnAIR-FD ($r=5$) and OnAIR-UD schemes, online inpainting with a fixed DCT dictionary, batch dictionary learning-based reconstruction ($r=5$), 2D (frame-by-frame cubic) interpolation, and 3D interpolation. The best PSNR for each undersampling is in bold.}
\label{tab:inpaint:coastguard:noisy}
\end{table}

\begin{figure*}[t!]
\begin{center}
\begin{tabular}{c}
\includegraphics[width=0.95\textwidth]{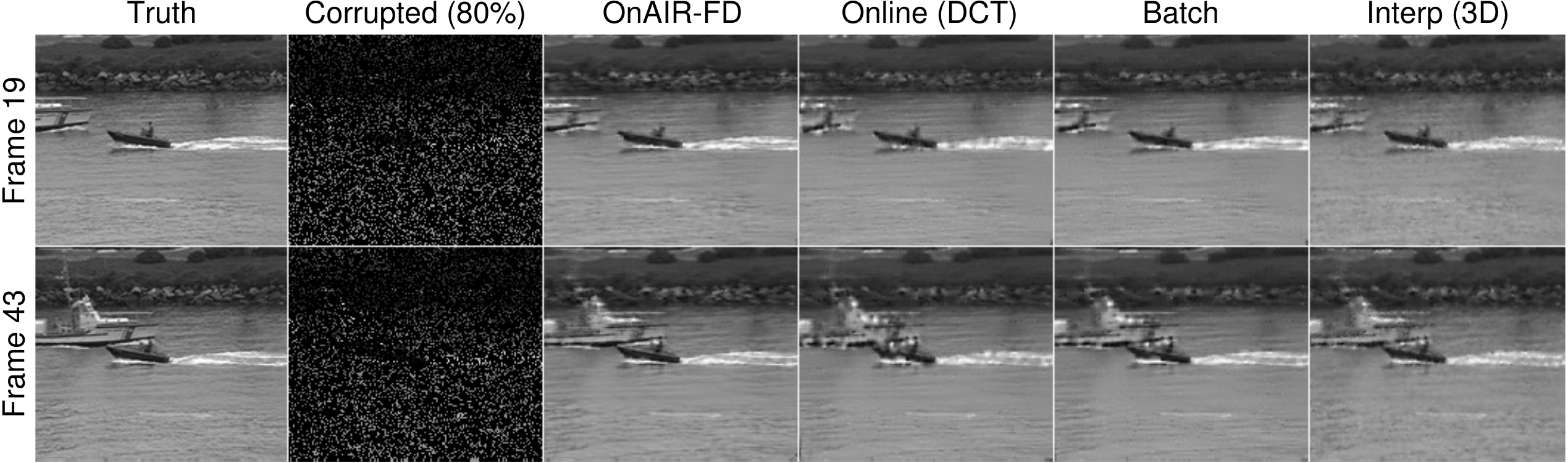}
\vspace{0.2cm} \\
\includegraphics[width=0.95\textwidth]{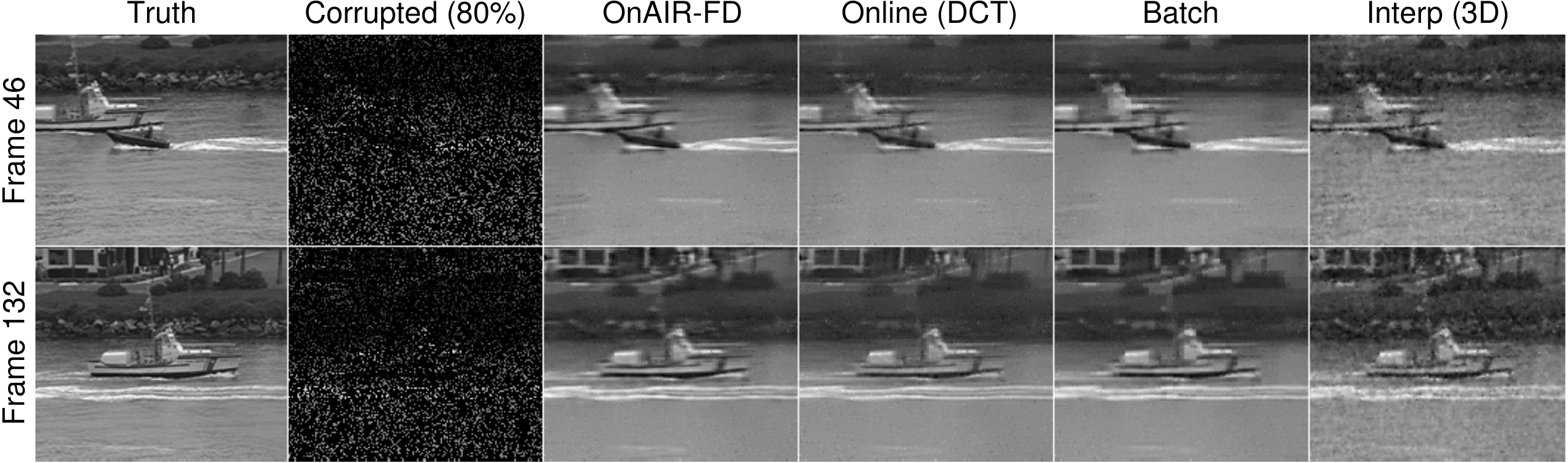}
\end{tabular}
\caption{Two representative frames from the reconstructions produced by various methods for the Coastguard video with 80\% missing pixels and no added noise (top), and 80\% missing pixels with 25 dB added Gaussian noise (bottom). Results are shown for the proposed OnAIR-FD ($r=5$) method, the online method with fixed DCT dictionary, the batch learning-based reconstruction method ($r=5$), and 3D interpolation. The true (reference) frames are also shown. Top: OnAIR-FD achieves PSNR (of video) improvements of 2.0 dB, 1.0 dB, and 1.4 dB respectively, over the aforementioned methods. Bottom: OnAIR-FD achieves PSNR improvements of 1.0 dB, 1.0 dB, and 1.5 dB respectively, over the aforementioned methods.}
\label{fig:inpaint:coastguard:recons}
\end{center}
\end{figure*}

\begin{figure*}[t!]
\begin{center}
\includegraphics[width=\textwidth]{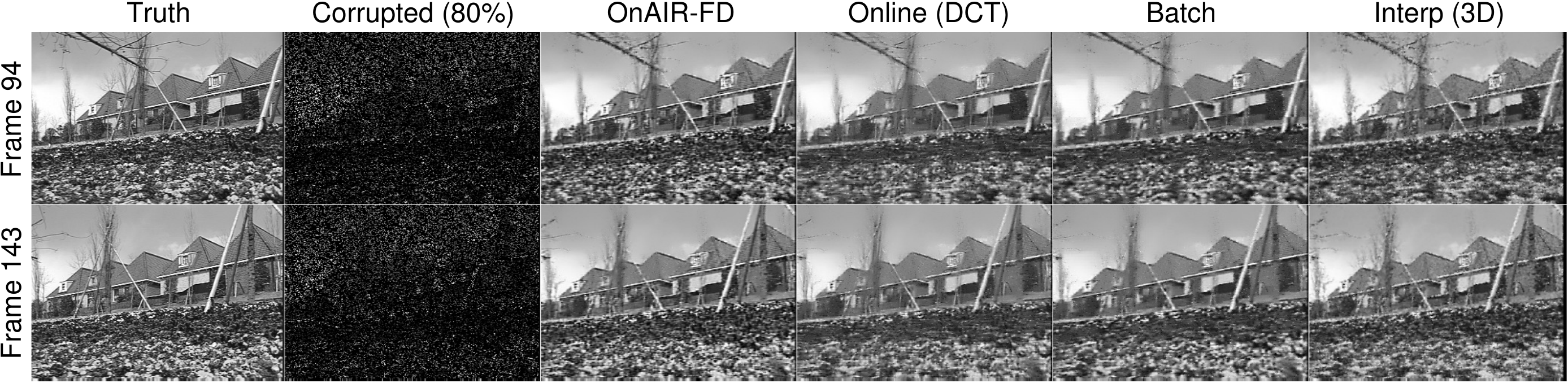}
\caption{Two representative frames from the reconstructions produced by various methods for the Flower Garden video with 80\% missing pixels. The methods considered are the proposed OnAIR-FD ($r=5$) method, the online method with fixed DCT dictionary, the batch learning-based reconstruction method, and 3D interpolation. The true frames are also shown. OnAIR-FD achieves PSNR (of video) improvements of 0.7 dB, 0.6 dB, and 1.3 dB respectively, over the aforementioned competing methods.}
\label{fig:inpaint:flower:recons}
\end{center}
\end{figure*}

\begin{figure}[t!]
\centering
\includegraphics[width=\linewidth]{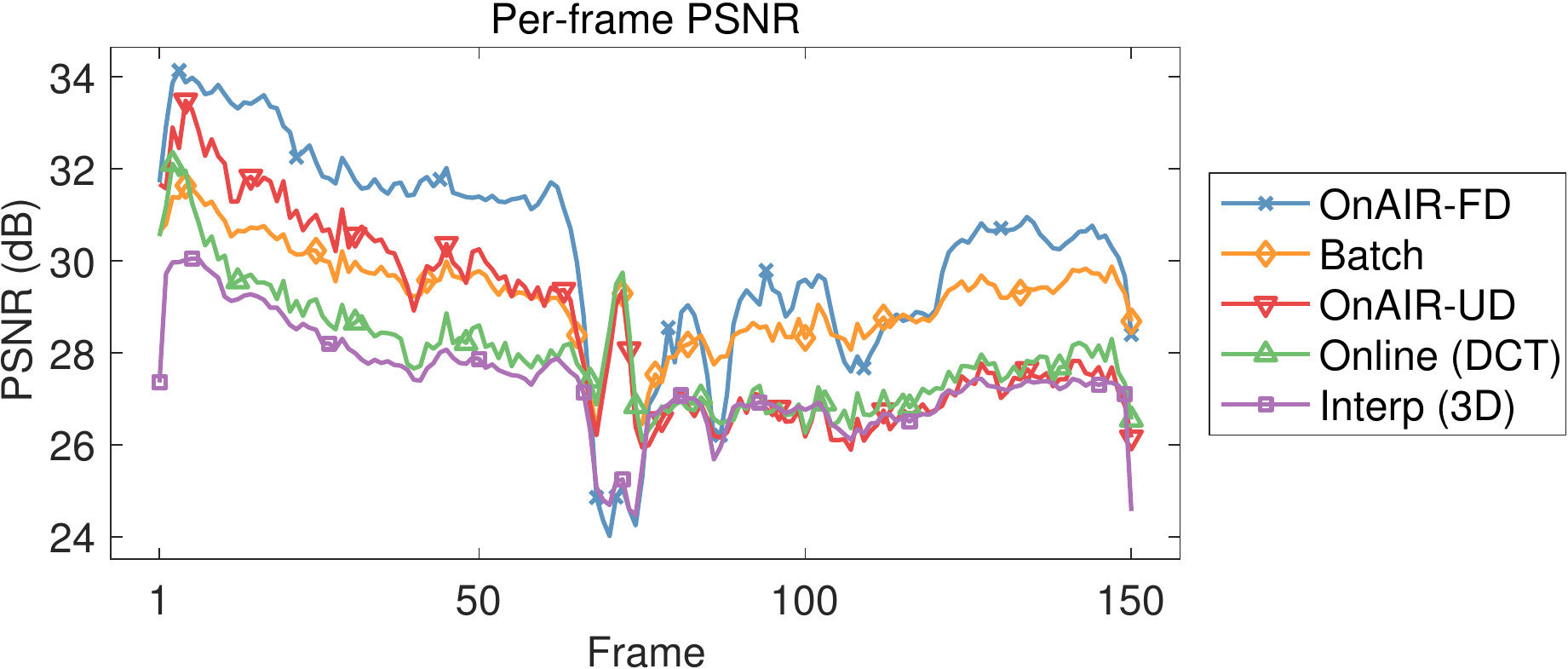}
\caption{Per-frame (2D) PSNR values (in dB) for the reconstructions produced by OnAIR-FD ($r=5$), OnAIR-UD, the online method with fixed DCT dictionary, the batch learning-based method, and 3D interpolation, for the Coastguard video with 70\% missing pixels.}
\label{fig:inpaint:psnrs:coastguard}
\end{figure}

%
%
%
%
%
%
%

\subsubsection{Results}
Table~\ref{tab:inpaint} lists the PSNR values (in dB) for the various reconstruction methods for different levels of subsampling (from 50\% to 90\% missing pixels) of three videos (without added noise) from the BM4D dataset. 
Table~\ref{tab:inpaint:coastguard:noisy} shows analogous results for the Coastguard video when i.i.d. zero-mean Gaussian noise with 25 dB PSNR was added before sampling the pixels. The proposed OnAIR-FD method typically provides the best PSNRs at higher undersampling factors, and the proposed OnAIR-UD method, which uses a more structured (unitary) dictionary\footnote{OnAIR-UD can be interpreted as a unitary sparsifying transform model~\cite{sabres} with $D^{H}$ denoting the sparsifying transform.}, performs better at lower undersampling factors.

The best PSNRs achieved by the OnAIR methods are typically better (by up to 1.4 dB) than for the Batch Learning scheme. Both OnAIR variations also typically outperform (by up to 2 dB) the Online DCT method, which uses a fixed DCT dictionary---the initial $D$ in our methods.

Figs.~\ref{fig:inpaint:coastguard:recons} and \ref{fig:inpaint:flower:recons} show the original and reconstructed frames for a few representative frames of the Coastguard and Flower Garden videos. Results for multiple methods are shown.
Fig.~\ref{fig:inpaint:coastguard:recons} shows that the proposed OnAIR-FD method produces visually more accurate reconstructions of the texture in the waves and also produces fewer artifacts near the boats in the water and the rocks on the shore.
Fig.~\ref{fig:inpaint:flower:recons} illustrates that the proposed OnAIR method produces a sharper reconstruction with less smoothing artifacts than the online method with a fixed DCT dictionary and the batch learning-based reconstruction method, and it is less noisy than the interpolation-based reconstruction.

\subsubsection{Properties}
Fig.~\ref{fig:inpaint:psnrs:coastguard} shows the frame-by-frame (2D) PSNRs for the Coastguard video inpainted from 70\% missing pixels, using various methods. Clearly the proposed OnAIR-FD method achieves generally higher PSNRs across frames of the video. The decrease in PSNR between frames 70 and 80 in the video is due to the significant motion in the scene that occurs in these frames. This change in performance is typical for dictionary learning-based methods because the sparse codes for the corresponding image patches and dictionary require a few iterations to adapt to the scene. However, as illustrated in Fig.~\ref{fig:inpaint:coastguard:recons}, the OnAIR methods may produce visually superior reconstructions in such cases.

%
%
%
%
%
%

To assess the relative efficiency of the OnAIR methods compared to the Batch Learning method, we measured runtimes on the Coastguard video with the same fixed patch sizes, patch strides, and numbers of iterations for each method. The experiments were performed in MATLAB R2017b on a 2016 MacBook Pro with a 2.7 GHz Intel Core i7 processor and 16 GB RAM. With these settings, the OnAIR-FD and OnAIR-UD methods required an average of 4.4 seconds and 1.9 seconds to process each frame, respectively, while the Batch Learning method required an average of 7.9 seconds. Thus the OnAIR-LD and OnAIR-UD methods were 1.8x and 4.2x faster, respectively, than the Batch Learning method. In addition, the OnAIR methods typically require much fewer iterations (per minibatch) compared to the batch method to achieve their best reconstruction accuracies, so, in practice, when utilizing half as many outer iterations, the OnAIR-FD and OnAIR-UD methods were approximately 3.6x and 8.0x faster, respectively, than the batch scheme.

The decreased memory requirement of the proposed OnAIR method compared to batch learning methods such as from \cite{sravbrijefraj} is often crucial in practice. Indeed, even on the modestly sized Coastguard dataset, which contains 150 frames of resolution $144 \times 172$, constructing the patch matrix and other associated matrices necessary to run the Batch Learning method with patch strides of $1 \times 1 \times 1$ or $2 \times 2 \times 1$ would have required more than the 16GB of RAM available on our computer (as a result, we chose a $4 \times 4 \times 1$ stride). Moreover, the memory requirement of the Batch Learning method scales linearly with the number of frames (since all frames are processed concurrently), so increasing the length of the frame sequence to, say, 300 frames would again necessitate more RAM than is available on typical machines. Conversely, the OnAIR method has a small, fixed memory footprint that is independent of the number of frames in the dataset (it depends instead on the number of frames \textit{per minibatch}) and thus can process datasets of arbitrary length.


\subsection{Dynamic MRI Reconstruction} \label{subsec:dmri}

\subsubsection{Framework}
Here, we demonstrate the usefulness of the proposed OnAIR methods for reconstructing dynamic MRI data from highly undersampled k-t space measurements. We work with the multi-coil (12-element coil array) cardiac perfusion data \cite{ota1} and the PINCAT data \cite{lingal16,shari12} from prior works. For the cardiac perfusion data, we retrospectively undersampled the k-t space using variable-density random Cartesian undersampling with a different undersampling pattern for each time frame, and for the PINCAT data we used pseudo-radial sampling with a random rotation of radial lines between frames. The undersampling factors tested are shown in Table~\ref{tab:mri}. For each dataset, we obtained reconstructions using the proposed OnAIR methods, the online method but with a fixed DCT dictionary, and the batch learning-based reconstruction (based on (P1)) method \cite{saibriarajjeff2}. We also ran the recent L+S \cite{ota1} and k-t SLR \cite{lingala2011accelerated} methods, two sophisticated batch methods for dynamic MRI reconstruction that process all the frames jointly. Finally, we also computed a baseline reconstruction in each experiment by performing zeroth order interpolation across time at non-sampled k-t space locations (by filling such locations with the nearest non-zero entry along time) and then backpropagating the filled k-t space to image space by pre-multiplying with the $A^{H}$ corresponding to fully sampled data. The first estimates ($f_{0}^{t}$) of newly arrived frames in our OnAIR methods are also computed via such zeroth order interpolation, but using only the estimated (i.e., already once processed and reconstructed) nearest (older) frames.

\begin{table*}[t!]
\resizebox{\textwidth}{!}{
\centering
\begin{tabular}{|c||c|c|c|c|c|c||c|c|c|c|c|c|}
\hline
\multirow{2}{*}{Acceleration} & \multicolumn{6}{c||}{Cardiac Perfusion} & \multicolumn{6}{c|}{PINCAT} \\
\cline{2-13}
                   & 4x & 8x & 12x & 16x & 20x & 24x & 5x & 6x & 7x & 9x & 14x & 27x \\
\hline \hline
OnAIR-LD & \textbf{10.2\%} & \textbf{12.8\%} & \textbf{14.8\%} & \textbf{16.7\%} & \textbf{18.1\%} & \textbf{18.0\%} & \textbf{8.9\%} & \textbf{9.7\%} & \textbf{11.0\%} & \textbf{12.4\%} & \textbf{15.5\%} & 21.8\% \\ \hline
Online (DCT)    & 10.8\% & 13.7\% & 15.8\% & 18.2\% & 20.7\% & 20.8\% & 9.5\% & 10.2\% & 11.5\% & 13.2\% & 16.4\% & 22.5\% \\ \hline
Batch Learning  & 10.7\% & 13.7\% & 15.9\% & 18.2\% & 22.0\% & 23.9\% & 10.0\% & 10.7\% & 11.8\% & 13.2\% & 15.9\% & \textbf{20.9\%} \\ \hline
L+S             & 11.0\% & 13.8\% & 16.1\% & 18.4\% & 21.5\% & 22.5\% & 11.8\% & 12.9\% & 14.4\% & 16.6\% & 20.0\% & 25.9\% \\ \hline
k-t SLR         & 11.2\% & 15.7\% & 18.4\% & 21.3\% & 24.3\% & 26.5\% & 9.8\% & 10.9\% & 12.4\% & 14.7\% & 18.2\% & 24.2\% \\ \hline
Baseline        & 12.8\% & 15.9\% & 18.9\% & 21.1\% & 24.5\% & 28.1\% & 22.3\% & 24.7\% & 27.5\% & 31.3\% & 36.6\% & 44.5\% \\ \hline
\end{tabular}
}
\caption{NRMSE values as percentages at several undersampling factors for the cardiac perfusion data with Cartesian sampling (left) and for the PINCAT data with pseudo-radial sampling (right). The methods compared are the proposed OnAIR-LD ($r=1$) method, the online scheme with a fixed DCT dictionary, the batch learning-based reconstruction ($r=1$), the L+S method, the k-t SLR method, and a baseline reconstruction. The best NRMSE achieved for each undersampling and each dataset is in bold.}
\label{tab:mri}
\end{table*}

For the online schemes, we used $8 \times 8 \times 5$ spatiotemporal patches with $\tilde{M}=5$ frames per temporal window (minibatch) and a (temporal) window stride of 1 frame. We extracted overlapping patches in each minibatch using a spatial stride of 2 pixels along each dimension. We learned a square $320 \times 320$ dictionary whose atoms when reshaped into $64 \times 5$ space-time matrices had rank $r = 1$, which worked well in our experiments. A forgetting factor of $\rho = 0.9$ was observed to work well.
We ran the online schemes for $K=7$ outer iterations per minibatch, with $\hat{K}=1$ iteration in the dictionary learning step and $\tilde{K} = 10$ proximal gradient steps in the image update step, respectively. Prior to the first outer iteration for each minibatch in (P1), the sparse coefficients were updated using $3$ block coordinate descent iterations to allow better adaptation to new patches. We used $K=50$ outer iterations for the first minibatch to warm start the algorithm and used an initial DCT $\hat{D}^{0}$, and the initial sparse codes were zero.
After one complete pass of the online schemes over all the frames, we performed another pass over the frames, using the reconstructed frames and learned dictionary from the first pass as the first initializations in the second pass. The additional pass gives a slight image quality improvement over the first (a single pass corresponds to a fully online scheme) as will be shown later.

For the batch dictionary learning-based reconstruction method, we used the same patch dimensions, strides, and initializations
as for the online methods. We ran the batch method for 50 iterations. For the L+S and k-t SLR methods, we used the publicly available MATLAB implementations from \cite{ota2} and \cite{lingala2}, respectively, and ran each method to convergence. The regularization parameters (weights) for all the methods here were tuned for each dataset by sweeping them over a range of values and selecting values that achieved good reconstruction quality at intermediate k-t space undersampling factors. We measured the dMRI reconstruction quality using the normalized root mean square error (NRMSE) metric expressed as percentage that is computed as
\begin{equation} \label{eq:nrmse}
\mathrm{NRMSE}(\hat{x}) = \dfrac{\|\hat{x} - x_{\text{ref}}\|_2}{\|x_{\text{ref}}\|_2} \times 100\%,
\end{equation}
where $\hat{x}$ is a candidate reconstruction and $x_{\text{ref}}$ is the reference reconstruction (e.g., computed from ``fully" sampled data).


\begin{figure*}[t!]
\centering
\begin{subfigure}[t]{0.48\linewidth}
\centering
\includegraphics[width=\linewidth]{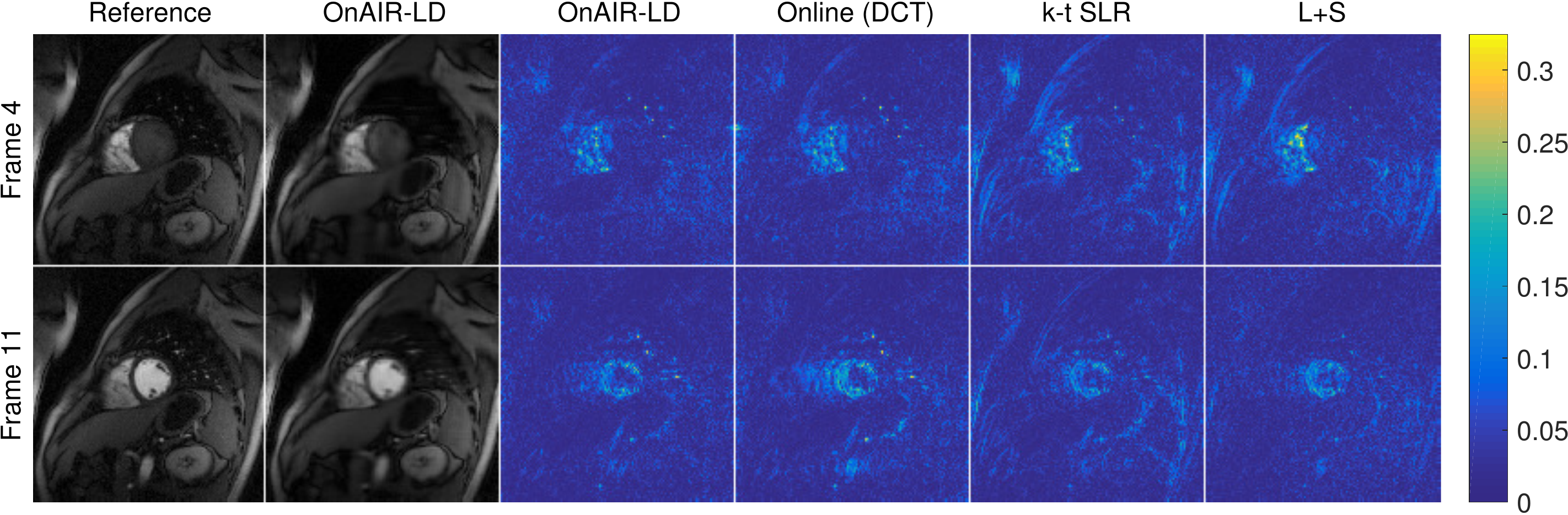}
\end{subfigure}%
\hspace{0.5cm}
\begin{subfigure}[t]{0.48\linewidth}
\centering
\includegraphics[width=\linewidth]{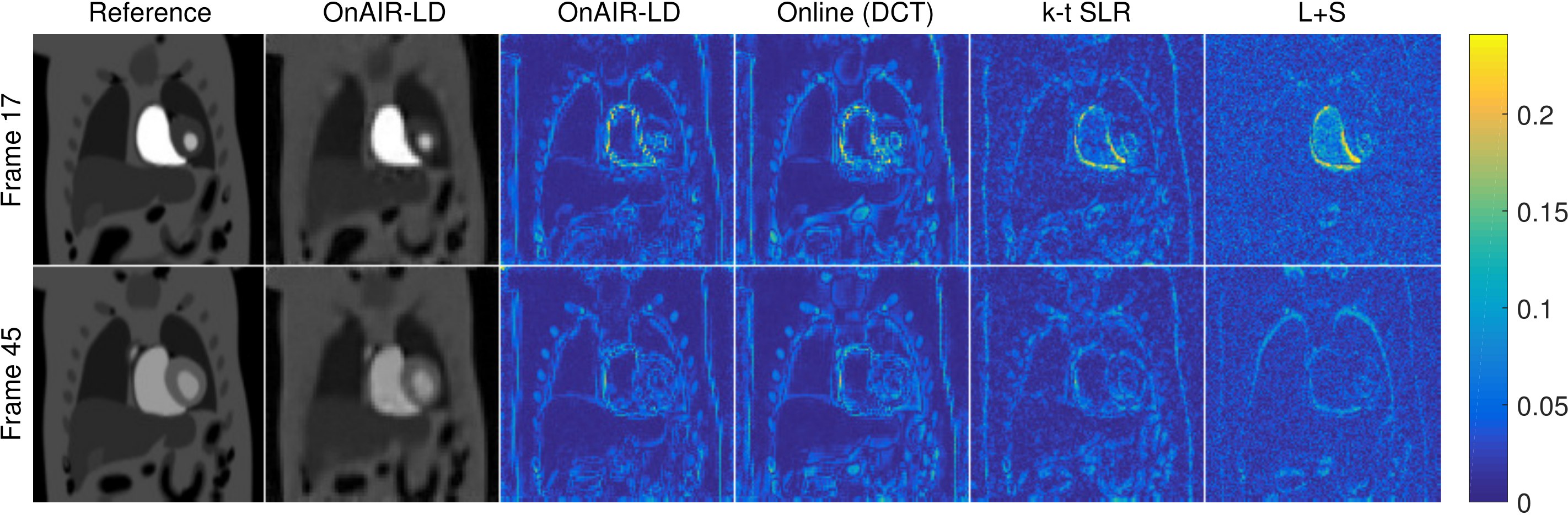}
\end{subfigure}
\caption{Left: reconstructions of the cardiac perfusion data with 12x undersampling (Cartesian sampling). Right: reconstructions of the PINCAT data with 7x undersampling (pseudo-radial sampling). Each panel shows two representative frames from a reference (fully sampled) reconstruction along with the corresponding frames from the proposed OnAIR-LD ($r=1$) scheme. The right four columns of each panel depict the corresponding reconstruction error magnitudes (w.r.t. reference) for OnAIR-LD, the online method with a fixed DCT dictionary, the k-t SLR method, and the L+S method, respectively. Compared to the competing methods, OnAIR-LD achieves (3D) NRMSE improvements of 0.6 dB, 1.9 dB, and 0.7 dB, respectively, and 0.4 dB, 1.0 dB, and 2.3 dB, respectively, for the cardiac perfusion and PINCAT data.}
\label{fig:mri:recons}
\end{figure*}

\begin{figure}[!t]
\begin{center}
\includegraphics[width=0.9\linewidth]{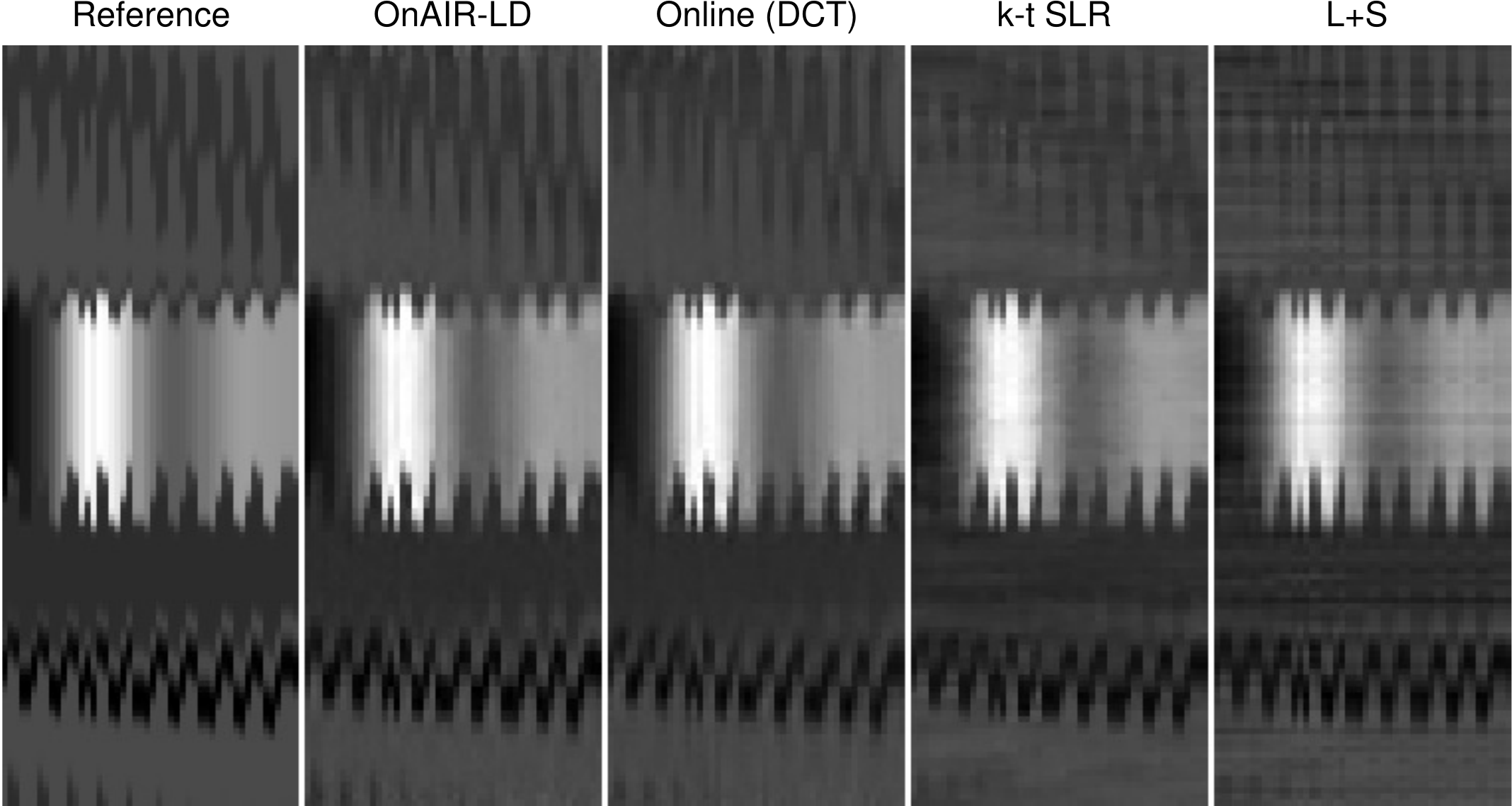}
\caption{Temporal ($y−-t$) profiles of a spatial vertical line cross section for the reference PINCAT reconstruction, the proposed OnAIR-LD ($r=1$) method, the online method with fixed DCT dictionary, the k-t SLR method, and the L+S method for 14x undersampling (pseudo-radial sampling).}
\label{fig:mri:pincat:yt}
\end{center}
\end{figure}

\begin{table}[!t]
\resizebox{\linewidth}{!}{
\centering
\begin{tabular}{|c|c|c|c|c|c|c|}
\hline
Acceleration               & 4x & 8x & 12x & 16x & 20x & 24x \\
\hline \hline
Online (oracle)            & \textbf{10.2\%} & \textbf{12.4\%} & \textbf{14.4\%} & \textbf{16.4\%} & \textbf{17.8\%} & \textbf{17.9\%} \\ \hline 
OnAIR-LD (2 passes) & \textbf{10.2\%} & 12.8\% & 14.8\% & 16.7\% & 18.1\% & 18.0\% \\ \hline 
OnAIR-LD (1 pass)   & \textbf{10.2\%} & 12.9\% & 14.8\% & 16.6\% & 18.3\% & 18.1\% \\ \hline 
OnAIR-UD           & 10.5\% & 13.6\% & 15.7\% & 17.8\% & 20.4\% & 20.1\% \\ \hline 
Online (DCT)               & 10.8\% & 13.7\% & 15.8\% & 18.2\% & 20.7\% & 20.8\% \\ \hline 
\end{tabular}}
\caption{NRMSE values as percentages for the cardiac perfusion data at seversal undersampling factors with Cartesian sampling. The methods compared are an oracle online scheme (where the dictionary is learned from patches of the reference and fixed during reconstruction), the proposed OnAIR-LD ($r=1$) method with a single pass or two passes over the frames, the proposed OnAIR-UD method, and the online scheme with a fixed DCT dictionary. The best NRMSE for each undersampling is in bold.}
\label{tab:mri:otazo:oracle}
\end{table}

\subsubsection{Results and Comparisons}
Table~\ref{tab:mri} shows the reconstruction NRMSE values obtained using various methods for the cardiac perfusion and PINCAT datasets at several undersampling factors.
The proposed OnAIR-LD method achieves lower NRMSE values in almost every case compared to the L+S and k-t SLR methods, the Online DCT scheme, the baseline reconstruction, and the batch learning-based reconstruction scheme. In particular, OnAIR-LD achieves a peak improvement of 2.5 dB compared to the Batch Learning method, and it achieves a peak improvement of 1.3 dB compared to the Online DCT method. Unlike the batch schemes (i.e., L+S, k-t SLR, and the batch learning-based reconstruction) that process or learn over all the data jointly, the OnAIR-LD scheme only directly processes data corresponding to a minibatch of $5$ frames at any time. Yet OnAIR-LD outperforms the other methods because of its memory (via the forgetting factor) and local temporal adaptivity (or tracking).
These results show that the proposed OnAIR methods are well-suited for processing streaming data.

%
%
%
%
%

Fig.~\ref{fig:mri:recons} shows the reconstructions and reconstruction error maps (magnitudes displayed) for some representative frames from the cardiac perfusion and PINCAT datasets at 12x and 7x undersampling, respectively. The error maps indicate that the proposed OnAIR-LD scheme often produces fewer artifacts compared to the existing methods.

Fig.~\ref{fig:mri:pincat:yt} shows the reconstructed $y-t$ profiles for various methods obtained by extracting the same vertical line segment from each reconstructed frame of the PINCAT data and concatenating them. The Online DCT scheme and the batch methods L+S and k-t SLR show line-like or additional smoothing artifacts that are not produced by the proposed OnAIR method, which suggests that the OnAIR method produces reconstructions with greater temporal resolution.

Note that we used full-rank ($r=5$) atoms in the video inpainting experiments, while in our dynamic MRI reconstruction experiments we chose low-rank ($r=1$) atoms. Intuitively, low-rank atoms are a better model for the dynamic MRI data because the videos have high temporal correlation and rank-$1$ atoms are necessarily constant in their temporal dimension. Conversely, the videos used in the inpainting experiments contained significant camera motion and thus dictionary atoms with more temporal variation (\ie higher rank) enabled more accurate reconstructions.

\subsubsection{Properties}
Table~\ref{tab:mri:otazo:oracle} investigates the properties of the proposed OnAIR methods in more detail using the cardiac perfusion data. Specifically, it compares the NRMSE values produced by the OnAIR-LD scheme with one or two passes over the data, the OnAIR-UD method, and the online method with a fixed DCT dictionary. 
In addition, we ran the online method but with a fixed ``oracle" dictionary learned from patches of the reference (true) reconstruction by solving the DINO-KAT learning problem \eqref{eq:dl:dinokat}. 
The oracle dictionary was computed based on the ``fully" sampled data, so it can be viewed as the ``best" dictionary that one could learn from the undersampled dataset. From Table~\ref{tab:mri:otazo:oracle}, we see that the NRMSE values achieved by the OnAIR-LD scheme with two passes are within 0.0\% - 0.5\% of the oracle NRMSE values, which suggests that the proposed scheme is able to learn dictionaries with good representational (recovery) qualities from highly undersampled data.
Moreover, the performance of the OnAIR-LD scheme with a single pass is almost identical to that with two passes, demonstrating the promise of fully online dMRI reconstruction.
The OnAIR-LD method outperformed the OnAIR-UD scheme for the cardiac perfusion data indicating that temporal low-rank properties better characterize the dataset than unitary models.
The OnAIR schemes achieved NRMSEs that were as much as 13.9\% lower than the nonadaptive Online DCT scheme at higher undersampling rates, which suggests that the learned dictionaries were able to better uncover the hidden structure of the scene.

%
%
%
%
%
%

To assess the relative efficiency of the OnAIR methods compared to the Batch Learning method, we measured runtimes on the PINCAT dataset with the same parameter settings for each method used to generate the results in Table~\ref{tab:mri}. With these settings, the OnAIR-LD and OnAIR-UD methods required an average of 8.5 seconds and 2.1 seconds to process each frame, respectively, while the Batch Learning method required an average of 84.0 seconds. Thus the OnAIR-LD and OnAIR-UD methods were 9.9x and 40.0x faster, respectively, than the Batch Learning method. One key reason for these improvements is that 50 outer iterations were required by the batch method while only 7 outer iterations were required by the online methods (after the first minibatch) to achieve the reported reconstruction accuracies; this result suggests that the online dictionary adaptation performed by the OnAIR methods is both computationally efficient and allows the model to better adapt to the underlying structure of the data. The additional speedup of the OnAIR-UD method with respect to OnAIR-LD is attributable to the relative efficiency of the matrix-valued $(D, Z)$ updates of OnAIR-UD compared to the less optimized block coordinate descent iterations over the columns of $D$ and $C$ prescribed by OnAIR-LD.


\section{Conclusions} \label{sec:conclusion}

This paper has presented a framework for online estimation of dynamic image sequences by learning dictionaries. Various properties were also studied for the learned dictionary such as a unitary property and low-rank atoms, which offer additional efficiency or robustness to artifacts. 
The proposed OnAIR algorithms sequentially and efficiently update the images, dictionary, and sparse coefficients of image patches from streaming measurements. 
Importantly, our algorithms can process arbitrarily long video sequences with low memory usage over time. Our numerical experiments demonstrated that the proposed methods produce accurate reconstructions for video inpainting and dynamic MRI reconstruction. The proposed methods may also be suitable for other inverse problems, including medical imaging applications such as interventional imaging, and other video-processing tasks from computer vision. We hope to investigate other application domains as well as study potential real-time applicability for OnAIR approaches in future work.

\bibliographystyle{IEEEtran}
\bibliography{paper}

\begin{IEEEbiography}[{\includegraphics[width=1in,height=1.25in,clip,keepaspectratio]{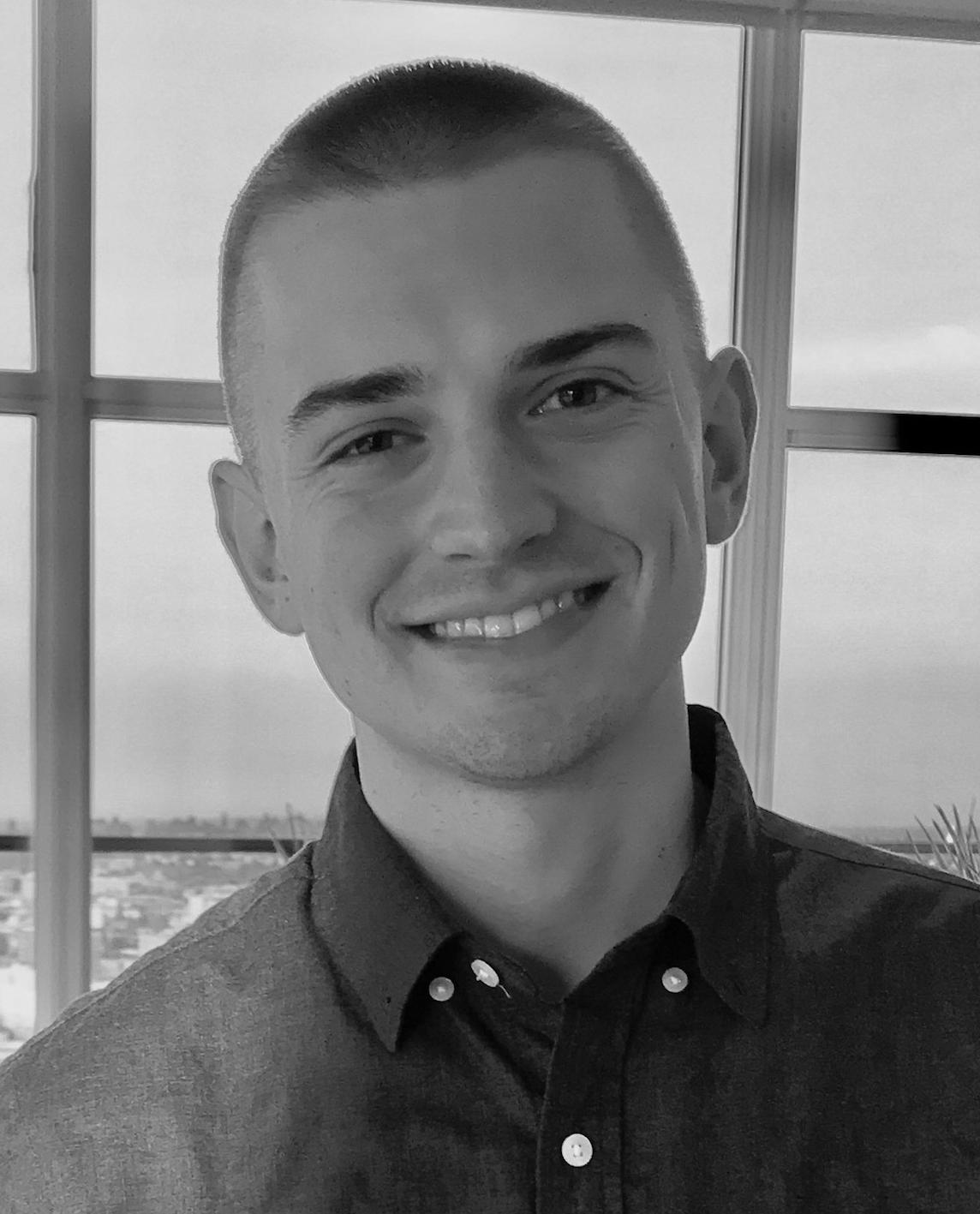}}]
{Brian E. Moore} received the B.S. degree in electrical engineering from Kansas State University, Manhattan, KS, USA, in 2012, and the M.S. and Ph.D. degrees in electrical engineering from the University of Michigan, Ann Arbor, MI, USA, in 2014 and 2018, respectively. He is the Co-Founder and CTO of Voxel51, Inc., Ann Arbor, MI, USA, a startup founded in late 2016 focused on cutting edge problems in computer vision and machine learning with applications in public safety and automotive sensing. His research interests include efficient algorithms for large-scale machine learning problems with a particular emphasis on computer vision applications.
\end{IEEEbiography}

\begin{IEEEbiography}[{\includegraphics[width=1in,height=1.25in,clip,keepaspectratio]{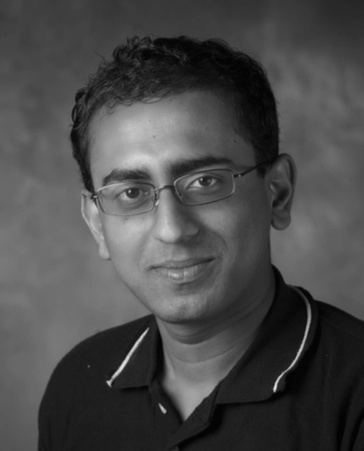}}]
{Saiprasad Ravishankar} received the B.Tech. degree in Electrical Engineering from the Indian Institute of Technology (IIT) Madras, India, in 2008, and the M.S. and Ph.D. degrees in Electrical and Computer Engineering in 2010 and 2014 respectively, from the University of Illinois at Urbana-Champaign. After his Ph.D., Dr. Ravishankar was an Adjunct Lecturer in the Department of Electrical and Computer Engineering, and a Postdoctoral Research Associate in the Coordinated Science Laboratory at the University of Illinois. From August 2015, he was a Research Fellow in the Department of Electrical Engineering and Computer Science at the University of Michigan, Ann Arbor. He was a Postdoc Research Associate in the Theoretical Division at Los Alamos National Laboratory from August 2018 to February 2019. He is currently an Assistant Professor in the Departments of Computational Mathematics, Science and Engineering, and Biomedical Engineering at Michigan State University. His research interests include signal and image processing, biomedical and computational imaging, data-driven methods, machine learning, signal modeling, inverse problems, data science, compressed sensing, and large-scale data processing. He was a recipient of the IEEE Signal Processing Society Young Author Best Paper Award for 2016. He is currently a member of the IEEE Computational Imaging Technical Committee.
\end{IEEEbiography}

\begin{IEEEbiography}[{\includegraphics[width=1in,height=1.25in,clip,keepaspectratio]{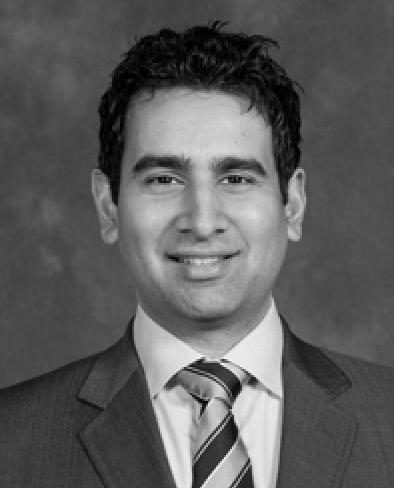}}]
{Raj Rao Nadakuditi} received the Ph.D. degree in 2007 from the Massachusetts Institute of Technology, Cambridge, MA, USA, and the Woods Hole Oceanographic Institution, Woods Hole, MA, USA. He is currently an Associate Professor in the Department of Electrical Engineering and Computer Science, University of Michigan, Ann Arbor, MI, USA. His research focuses on developing theory for random matrices for applications in signal processing, machine learning, queuing theory, and scattering theory. He received an Office of Naval Research Young Investigator Award in 2011, an Air Force Office of Scientific Research Young Investigator Award in 2012, the Signal Processing Society Young Author Best Paper Award in 2012, and the DARPA Young Faculty Award in 2014.
\end{IEEEbiography}

\begin{IEEEbiography}[{\includegraphics[width=1in,height=1.25in,clip,keepaspectratio]{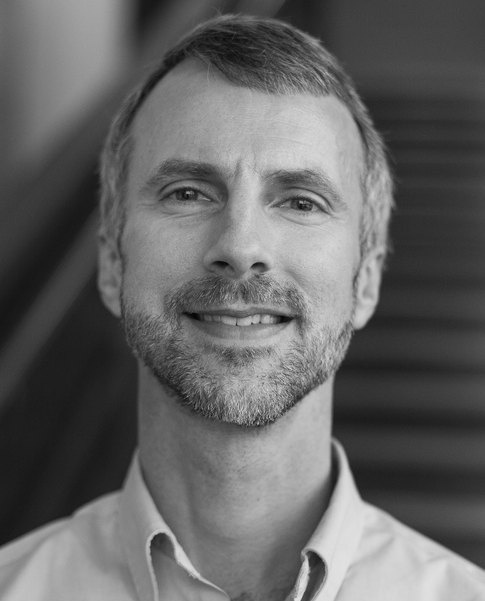}}]
{Jeffrey A. Fessler} is the William L. Root Professor of EECS at the University of Michigan. He received the BSEE degree from Purdue University in 1985, the MSEE degree from Stanford University in 1986, and the M.S. degree in Statistics from Stanford University in 1989. From 1985 to 1988 he was a National Science Foundation Graduate Fellow at Stanford, where he earned a Ph.D. in electrical engineering in 1990. He has worked at the University of Michigan since then. From 1991 to 1992 he was a Department of Energy Alexander Hollaender Post-Doctoral Fellow in the Division of Nuclear Medicine. From 1993 to 1995 he was an Assistant Professor in Nuclear Medicine and the Bioengineering Program. He is now a Professor in the Departments of Electrical Engineering and Computer Science, Radiology, and Biomedical Engineering. He became a Fellow of the IEEE in 2006, for contributions to the theory and practice of image reconstruction. He received the Francois Erbsmann award for his IPMI93 presentation, and the Edward Hoffman Medical Imaging Scientist Award in 2013. He has served as an associate editor for the IEEE Transactions on Medical Imaging, the IEEE Signal Processing Letters, the IEEE Transactions on Image Processing, the IEEE Transactions on Computational Imaging, and is currently serving as an associate editor for SIAM J. on Imaging Science. He has chaired the IEEE T-MI Steering Committee and the ISBI Steering Committee. He was co-chair of the 1997 SPIE conference on Image Reconstruction and Restoration, technical program co-chair of the 2002 IEEE International Symposium on Biomedical Imaging (ISBI), and general chair of ISBI 2007. His research interests are in statistical aspects of imaging problems, and he has supervised doctoral research in PET, SPECT, X-ray CT, MRI, and optical imaging problems.
\end{IEEEbiography}


\newpage
\clearpage


{
\twocolumn[
\begin{center}
\Huge Online Adaptive Image Reconstruction (OnAIR) Using Dictionary Models: Supplementary Material
\vspace{0.2in}
\end{center}]
}



\section{Additional Numerical Results} \label{sec:experiments:additional}
Here we provide numerical results that extend our investigation from Section V of our manuscript.

\subsection{Additional Inpainting Results}

\begin{figure}[t!]
\centering
\includegraphics[width=\linewidth]{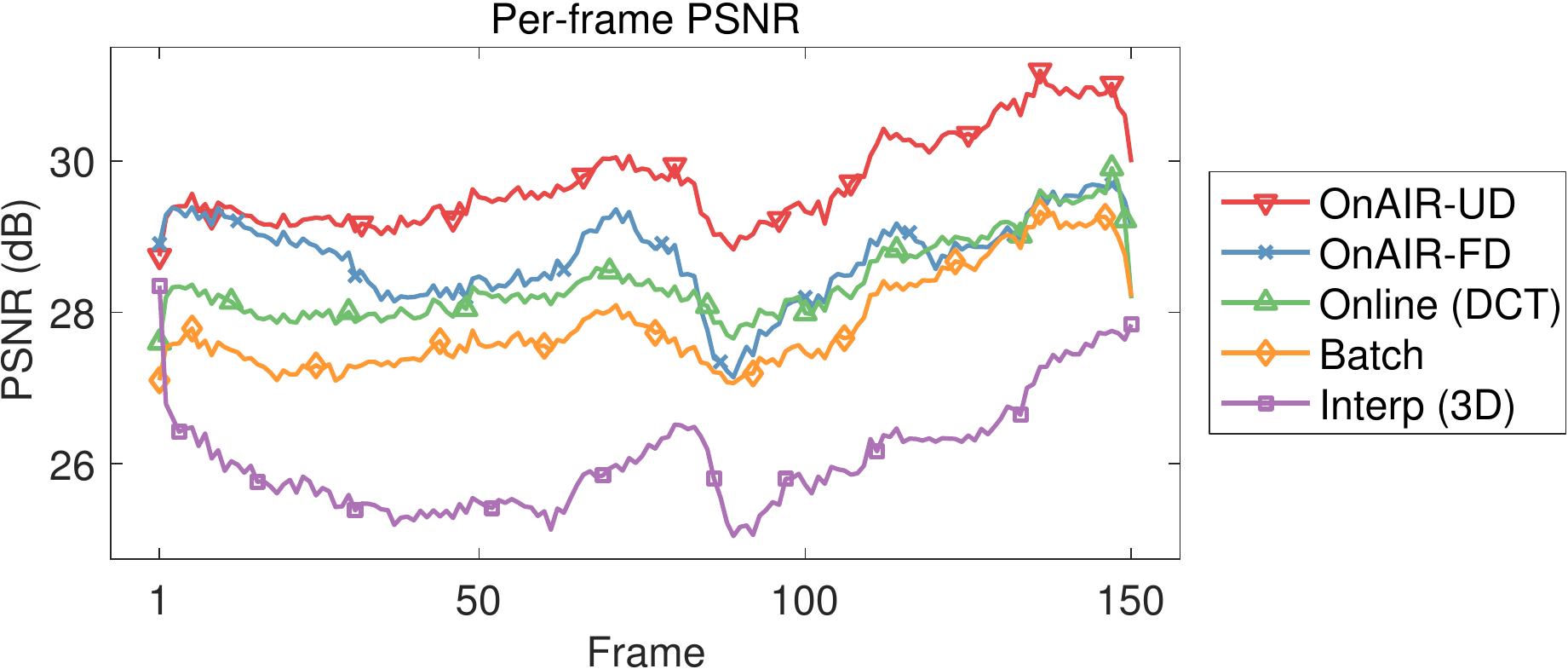}
\caption{Per-frame (2D) PSNR values (in dB) for the reconstructions produced by OnAIR-FD ($r=5$), OnAIR-UD, Online DCT, the Batch Learning method, and 3D interpolation, for the Bus video with 50\% missing pixels.}
\label{fig:inpaint:psnrs:bus}
\end{figure}

Fig.~\ref{fig:inpaint:psnrs:bus} shows the frame-by-frame (2D) PSNRs for the Bus video\footnote{The video was provided by the authors of the BM4D method \citeSupp{maggioni2013nonlocal:supp}. The dataset is publicly at \url{http://www.cs.tut.fi/~foi/GCF-BM3D}.} inpainted from 50\% missing pixels using various methods. Clearly the proposed OnAIR-UD scheme achieves consistently higher PSNRs across all frames. The overall trends in PSNR over frames are similar across the methods and are due to motion in the original videos, with more motion generally resulting in lower PSNRs.

\begin{figure}[t!]
\begin{center}
\includegraphics[width=\linewidth]{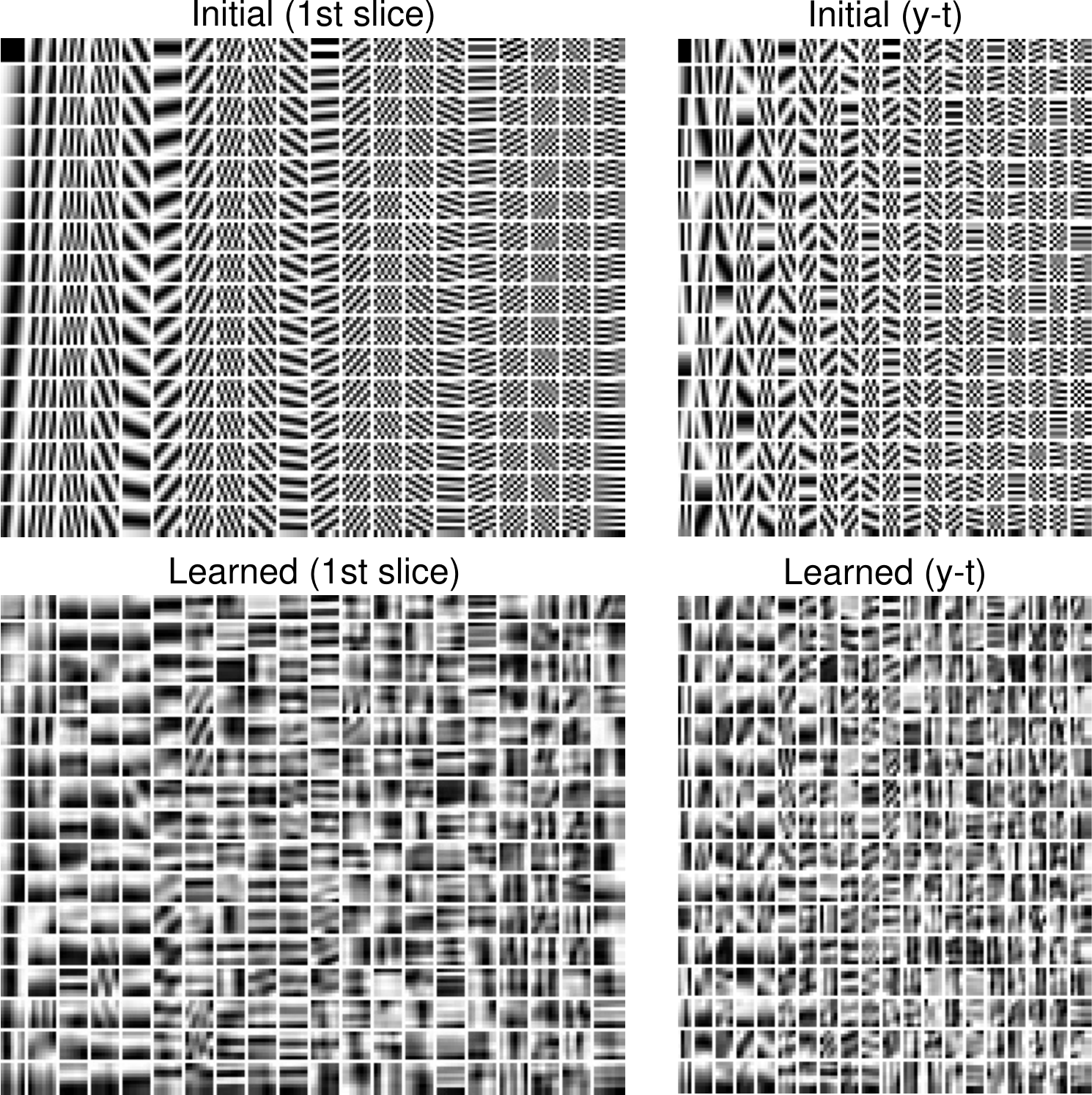}
\caption{Dictionaries for the Bus video with 50\% missing pixels. Top: the initial DCT dictionary. Bottom: the final learned dictionary produced by OnAIR-FD. Left: the first $8 \times 8$ ($x-y$) slice of each $8 \times 8 \times 5$ dictionary atom. Right: the $y-t$ profiles from a vertical cross-section through each $8 \times 8 \times 5$ atom.}
\label{fig:inpaint:bus:dict}
\end{center}
\end{figure}

Fig.~\ref{fig:inpaint:bus:dict} shows a representative example of a (final) learned dictionary produced by the proposed OnAIR-FD method for the Bus video, along with the initial (at $t=0$) DCT dictionary. The dictionaries each contain 320 atoms, each of which is an $8 \times 8 \times 5$ space-time tensor. We visualize each atom by plotting the first $8 \times 8$ $x-y$ slice of the atom and also plotting the $y-t$ profile from a vertical slice through the middle of each atom tensor. The $x-y$ (first slice) images show that the learned dictionary has adapted to both smooth and sharp gradients in the image, and the dynamic (evolved) nature of the $y-t$ profiles shows that the dictionary atoms have adapted to temporal trends in the data.

\subsection{Additional Dynamic MRI Reconstruction Results}

\begin{figure}[!t]
\begin{center}
\includegraphics[width=0.8\linewidth]{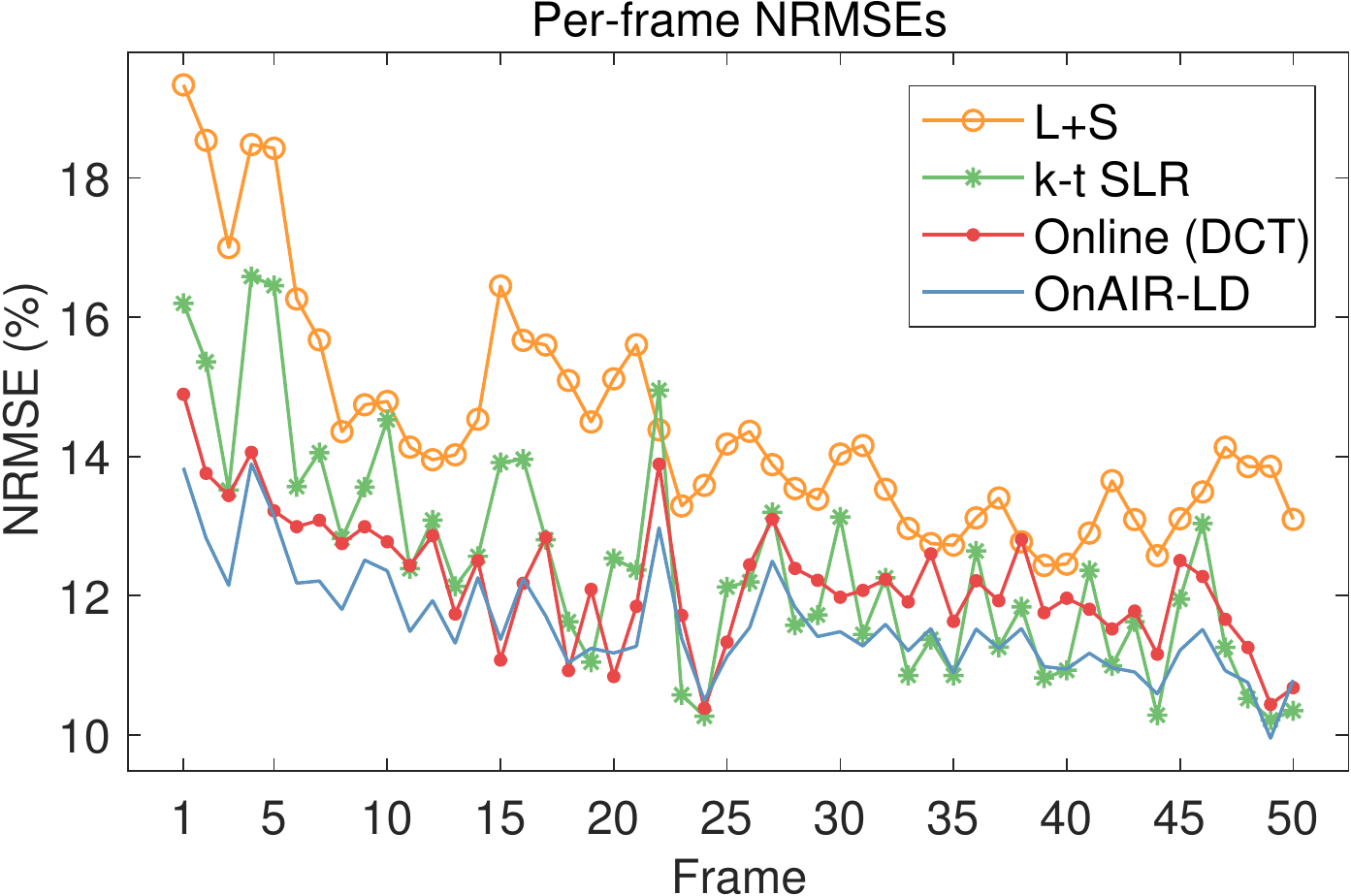}
\caption[]{Per-frame PSNR values for the reconstructions produced by OnAIR-LD ($r=1$), Online DCT, and the k-t SLR and L+S methods, for the PINCAT data from \citeSupp{lingal16:supp,shari12:supp} with 9x undersampling and pseudo-radial sampling.}
\label{fig:mri:pincat:psnrs}
\end{center}
\end{figure}

Fig.~\ref{fig:mri:pincat:psnrs} shows that the OnAIR-LD scheme typically achieves better frame-by-frame NRMSE compared to the other dynamic MRI reconstruction methods considered in this work. Note that such higher quality reconstructions are obtained in spite of the fact that the online scheme only processes and stores data corresponding to 5 frames (in $x^t$) at any time while the L+S, k-t SLR, and batch DINO-KAT methods process all data jointly.

\begin{figure}[!t]
\begin{center}
\includegraphics[width=\linewidth]{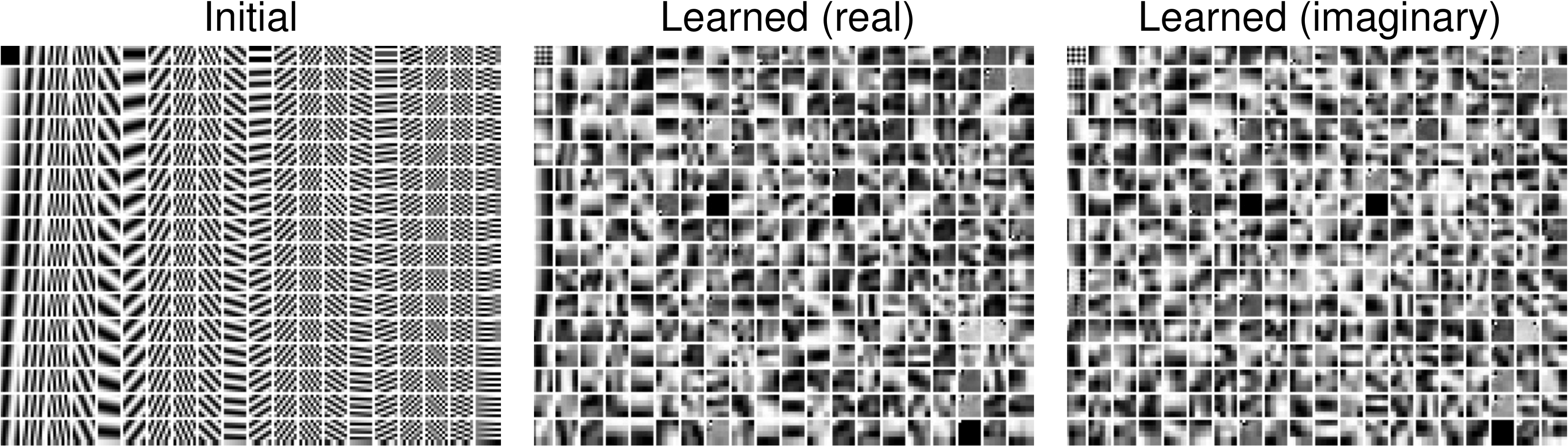}
\caption[]{Dictionaries for the PINCAT data from \citeSupp{lingal16:supp,shari12:supp} with 9x undersampling. Left: the atoms of the initial DCT dictionary. Center and right: the real and imaginary parts of the final learned dictionary produced by OnAIR-LD ($r=1$). The dictionary atoms are $8 \times 8 \times 5$ tensors, and only the first $8 \times 8$ ($x-y$) slice of each atom is displayed.}
\label{fig:mri:pincat:dict}
\end{center}
\end{figure}

Fig.~\ref{fig:mri:pincat:dict} shows an example of a learned dictionary produced by the proposed OnAIR-LD method on the PINCAT dataset at 9x undersampling, which is compared with the initial DCT dictionary. The dictionaries have $320$ atoms, each a $8 \times 8 \times 5$ complex-valued space-time tensor. Since the OnAIR LD dictionary atoms have rank $r=1$ when reshaped into $64 \times 5$ space-time matrices, we directly display the real and imaginary parts of the first $8 \times 8$ ($x-y$) slice of each learned atom. The initial DCT is also displayed similarly. 
The (eventual) learned dictionary has clearly evolved significantly from the initial DCT atoms and has adapted to certain smooth and sharp textures at various orientations in the data.

\bibliographystyleSupp{IEEEtran}
\bibliographySupp{paper}

\end{document}